\documentclass{article}

\usepackage{xcolor} 

\PassOptionsToPackage{numbers, compress}{natbib}
\usepackage[final]{neurips_2021}

\usepackage[utf8]{inputenc} 
\usepackage[T1]{fontenc}    
\usepackage{hyperref}       
\usepackage{url}            
\usepackage{booktabs}       
\usepackage{amsfonts}       
\usepackage{nicefrac}       
\usepackage{microtype}      
\usepackage{xcolor}         
\usepackage{wrapfig}
\usepackage{import}
\usepackage{subcaption}
\usepackage{tikz,pgfplots}
\usepackage{algorithm,algorithmicx,algpseudocode}

\usepackage{amsmath}
\usepackage{amsthm} 

\newcommand{\bx}{\mathbf{x}}

\newcommand{\bI}{\mathbf{I}}
\newcommand{\beps}{\boldsymbol\epsilon}
\newcommand{\bbeta}{\boldsymbol{\beta}}

\newcommand{\cN}{\mathcal{N}}
\newcommand{\cL}{\mathcal{L}}
\newcommand{\cF}{\mathcal{F}}

\newcommand{\real}{\mathbb{R}}
\newcommand{\zero}{\mathbf{0}}

\newcommand{\E}{\mathbb{E}}
\newcommand{\KL}{\mathbb{KL}}

\title{Bilateral Denoising Diffusion Models} 
\author{
  Max W. Y. Lam, Jun Wang, Rongjie Huang\thanks{Work done during an internship at Tencent AI Lab.}, Dan Su\\
  Tencent AI Lab\\
  Shenzhen, China\\
  \texttt{\{maxwylam, joinerwang\}@tencent.com}\\
  \And
  Dong Yu\\
  Tencent AI Lab\\
  Bellevue WA, USA\\
  \texttt{dyu@tencent.com}\\
}

\newtheorem{proposition}{Proposition}[section]

\newtheorem{remark}{Remark}[section]

\newtheorem{corollary}{Corollary}[section]

\newtheorem{note}{Note}[section]

\pgfplotsset{compat=1.17}
\begin{document}

\maketitle

\begin{abstract}
Denoising diffusion probabilistic models (DDPMs) have emerged as competitive generative models yet brought challenges to efficient sampling. In this paper, we propose novel bilateral denoising diffusion models (BDDMs), which take significantly fewer steps to generate high-quality samples. From a bilateral modeling objective, BDDMs parameterize the forward and reverse processes with a score network and a scheduling network, respectively. We show that a new lower bound tighter than the standard evidence lower bound can be derived as a surrogate objective for training the two networks. In particular, BDDMs are efficient, simple-to-train, and capable of further improving any pre-trained DDPM by optimizing the inference noise schedules. Our experiments demonstrated that BDDMs can generate high-fidelity samples with as few as 3 sampling steps and produce comparable or even higher quality samples than DDPMs using 1000 steps with only 16 sampling steps (a 62x speedup).
\end{abstract}

\section{Introduction}
Deep generative models have shown a tremendous advancement in image generation \cite{jonathan2020}, speech synthesis \cite{nanxin2020, zhifeng2021}, natural language generation \cite{Tom2020} and unsupervised representation learning \cite{Mark2020, jeff2019} for downstream tasks. The successful methods can be mainly divided into two branches -- generative adversarial network (GAN) \cite{ian2014} based models and likelihood-based models \cite{aaron2016, ali2019, laurent2016, diederik2018, tian2018, will2019, jonathan2019, george2021, sam2021, diederik2013, Danilo2014, lars2019, hinton2002,miguel2005}. The former use an adversarial training procedure, but the training can be unstable, and its training objective is not suitable to be compared against other GAN models; the latter use log-likelihood or surrogate objectives for training, but they also have intrinsic limitations. For example, the auto-regressive models \cite{aaron2016, ali2019} have exceedingly slow sampling speed and poor scaling properties on high dimensional data, as its sampling is inherently a sequential process. Likewise, the flow-based models \cite{laurent2016, diederik2018, tian2018, will2019, jonathan2019, george2021} rely on specialized architectures to build a normalized probability model so that the model training is less parameter-efficient and typically more difficult than other likelihood-based generative models \cite{sam2021}. On the other hand, for those using the surrogate losses, such as the evidence lower bound in variational auto-encoders \cite{diederik2013, Danilo2014, lars2019} and the contrastive divergence in energy-based models \cite{hinton2002,miguel2005}, they improve inference speed but often yield lower quality samples than autoregressive and GAN models and typically only work well for low-dimensional data \cite{sam2021}.

An up-and-coming class of likelihood-based models is the diffusion probabilistic models (DPMs) \cite{jonathan2020, jascha2015}, which shed light on the generation of high-quality samples comparable or even superior \cite{Prafulla2021} to the current state-of-the-art (SOTA) autoregressive and GAN models. Sohl-Dickstein et al. \cite{jascha2015} introduced the idea of using an iterative forward process for destroying the structure of a given distribution while learning the reverse process for restoring the structure in the data. With a different modeling objective, the score-based generative models \cite{yang2019} used a neural network trained with the score matching \cite{aapo2005} to produce samples via Langevin dynamics. As a seminal work of DPMs, Ho et al. \cite{jonathan2020} proposed the denoising diffusion probabilistic models (DDPMs), which exploited a connection \cite{pascal2011} between the above two ideas that the diffusion framework can be viewed as a specific approach to provide support to score matching. 
For the first time, DDPMs present high-quality image synthesis results on par with GAN models, especially on several higher-resolution image benchmarks.
Subsequently, score matching and denoising diffusion models were applied to speech synthesis, such as Wavegrad \cite{nanxin2020} and a parallel contribution Diffwave \cite{zhifeng2021}, and demonstrated the capability of generating high fidelity audio samples outperforming non-autoregressive adversarial models \cite{ryuichi2020,kundan2019, geng2020, Mikolaj2020} while matching SOTA autoregressive methods \cite{aaron2016, nal2018}. 

Despite the compelling results, the diffusion models are two to three orders of magnitude slower than other generative models such as GANs and VAEs. The major limitation of the generative diffusion models is that it requires up to thousands of diffusion steps during training to learn the target distribution; Therefore, a large number of denoising steps are often required at sampling time.
Recently, extensive investigations have been conducted to accelerate the sampling process for efficiently generating high-quality outputs. In WaveGrad \cite{nanxin2020}, a grid search algorithm was used to generate high-fidelity audio samples with six steps. From a different aspect, Song et al. \cite{yang2021} used a neural probability flow ODE to enable a fast deterministic multi-step sampling process. A parallel contribution called the denoising diffusion implicit models (DDIMs) \cite{jiaming2021} considered non-Markovian diffusion processes and used a subsequence of the noise schedule to accelerate the denoising process. A more recent work \cite{eric2021} explored a student-teacher method to distill the DDIMs sampling process into a single-step model.
 
The prior diffusion models only considered alternative reverse process forms or using additional knowledge for better conditional modeling. Distinctively, we noticed that the reduction of the sampling steps essentially depended on the choice of a noise schedule, which was conventionally considered yet as predetermined for the forward process.
We are thus motivated to learn the noise schedule within the network training framework directly. With such an incentive, we propose the bilateral denoising diffusion models (BDDM), which parameterize the forward and reverse processes, with a score network and a scheduling network, respectively. There are several nice properties of BDDMs. First, in BDDMs, we prove that a new lower bound for the log marginal likelihood tighter than the conventional ELBO can be derived. Secondly, we can also derive new objectives for learning both the score network and the scheduling network. Interestingly, our derived loss for the score network resembles the objective used in DDPMs under a reasonable condition. Thirdly, BDDMs allow for efficient training that merely adds a fraction of the original DDPM's score network training time. In particular, BDDM is efficient, simple to train, and can also be applied to any pre-trained DDPM for optimizing the noise schedules. Our experiments demonstrated that BDDMs can generate high-fidelity samples with as few as 3 sampling steps, and produce comparable or even higher quality samples than the SOTA DDPMs using 1000 steps with only 16 sampling steps (a 62x speedup).

\section{Diffusion probabilistic models (DPMs)}
Given i.i.d. samples $\{\bx_{0}\in\real^D\}$ from an unknown data distribution $p_\text{data}(\bx_0)$, diffusion probabilistic models (DPMs) \cite{jascha2015} define a forward process $q(\bx_{1:T}|\bx_0)=\prod_{t=1}^{T}q(\bx_{t}|\bx_{t-1})$ that converts any complex data distribution into a simple, tractable distribution after $T$ steps of diffusion. To revert the forward diffusion process, a reverse generative process, defined as a finite-time Markov chain $p_\theta(\bx_{t-1}|\bx_{t})$, is used to model the data distribution by the marginal likelihood $p_\theta(\bx_{0})=\int \prod_{t=1}^T p_\theta(\bx_{t-1}|\bx_{t}) d\bx_{1:T}$, where the variational parameters $\theta$ are learnt by maximizing the standard evidence lower bound (ELBO):
\begin{align}
\label{eq:elbo}
    \mathcal{F}_\text{elbo}
    :=
    \mathbb{E}_{q}
    \left[
    \log
    p_\theta(\mathbf{x}_0|\mathbf{x}_1)
    -
    \sum_{t=2}^{T}
    \mathbb{KL}\left(q(\mathbf{x}_{t-1}|\mathbf{x}_t,\mathbf{x}_0)||p_\theta(\mathbf{x}_{t-1}|\mathbf{x}_t)\right)
    -
    \mathbb{KL}\left(q(\mathbf{x}_{T}|\mathbf{x}_0)||p(\mathbf{x}_{T})\right)
    \right].
\end{align}

\subsection{Denoising diffusion probabilistic models (DDPMs)}
\label{sec:ddpm}
As an extension to DPMs, de-noising diffusion probabilistic models (DDPMs) \cite{jonathan2020} applied the score matching \cite{aapo2005, yang2019} technique to parameterize generative process. In particular, DDPMs considered a Gaussian diffusion process parameterized by a \textit{noise schedule} $\bbeta\in\real^T$ with $0<\beta_1, \dotsc, \beta_T<1$:
\begin{align}
    q_{\bbeta} (\bx_{1:T}|\bx_0) := \prod_{t=1}^{T} q_{\beta_t}(\bx_{t}|\bx_{t-1}),\qquad \text{where} \qquad q_{\beta_t}(\bx_{t}|\bx_{t-1}):= \cN(\sqrt{1-\beta_t}\bx_{t-1}, \beta_t\bI),
\end{align}
by which we can take advantage of a nice property of isotropic Gaussian distributions to directly express $\bx_t$ in a closed form:
\begin{align}
    q_{\bbeta}(\bx_t|\bx_0) = \cN\left(\bx_{t};\alpha_t\bx_{0},\left(1-\alpha_t^2\right) \bI\right),\qquad \text{where} \qquad \alpha_t=\prod_{i=1}^{t}\sqrt{1-\beta_i}
\end{align}
Then, a score network\footnote{Here, $\beps_{\theta}(\bx_{t},\alpha_t)$ is conditioned on the continuous noise scale $\alpha_{t}$, as in \cite{yang2021, nanxin2020}. Alternatively, the score network can also be conditioned on a discrete time index $\beps_{\theta}(\bx_{t},t)$, as in \cite{jiaming2021, jonathan2020}. An approximate mapping of a noise schedule to a time schedule \cite{kong2021fast} exists, we consider conditioning on noise scales as the general case.} $\beps_\theta (\bx_t, \alpha_t)$ is employed to define a more complex reverse process:
\begin{align}
    \label{eq:reverse}
   p_\theta(\bx_{t-1}|\bx_{t})
     &:=
    \cN
    \left(
        \frac
        {1}{\sqrt{1-\beta_{t}}}
        \left(
        \bx_{t}
        -
        \frac
        {\beta_{t}}
        {\sqrt{1-\alpha_{t}^2}}
        \beps_\theta\left(\bx_{t}, \alpha_{t}\right)
        \right),
        \frac
        {1-\alpha_{t-1}^2}
        {1-\alpha_{t}^2}\beta_{t}
        \bI
    \right),
\end{align}
in which case learning $\theta$ entails training the score network with back-propagation. Note that using the complete ELBO in Eq. (\ref{eq:elbo}) to train the score network requires $T$ forward passes of the score network and back-propagating through $T$ diffusion steps, which makes the training computationally prohibitive. To feasibly training score network, instead of computing the complete ELBO, DDPMs proposed a training heuristic to sample $t\sim \text{Uniform}(\{1,...,T\})$ at each step of training and compute the following simplified training loss:
\begin{align}
\label{eq:lddpm}
    \cL_\text{ddpm}^{(t)}(\theta)
    :=\left\| \beps_{t} - \beps_\theta\left(\alpha_{t}\bx_{0}+\sqrt{1-\alpha_t^2}\beps_{t}, \alpha_{t}\right) \right\|^2_2,
\end{align}
which can be seen as a re-weighted term of $\mathbb{KL}\left(q_{\bbeta}(\mathbf{x}_{t-1}|\mathbf{x}_t,\mathbf{x}_0)||p_\theta(\mathbf{x}_{t-1}|\mathbf{x}_t)\right)$. Notably, although the re-weighting effectively worked in practice, it is deficient for learning the noise schedule $\bbeta$. Distinctively, in the following section, we present an effective, theoretically grounded approach to estimate a proper inference schedule from an alternative bilateral modeling perspective.

\section{Bilateral denoising diffusion models (BDDMs)}
\label{gen_inst}
Noise scheduling has been shown significant \cite{nanxin2020, zhifeng2021, kong2021fast} for sampling acceleration \cite{nanxin2020} and for high-fidelity generation. In DDPMs, a linear noise schedule was set and shared for both forward and reverse processes \cite{jonathan2019}. In practice, a large $T$, such as $T=1000$, needs to be used to make the reverse process a good approximation. This leads to an unacceptably slow sampling process due to $T$ forward passes of the score network according to Eq. (\ref{eq:reverse}).

To relax the computational burden, Chen et al. \cite{nanxin2020} and Kong et al. \cite{zhifeng2021} allowed the noise schedule for sampling, denoted as $\hat{\bbeta}\in\real^N$, to be different from the noise schedule for training $\bbeta$. In particular, much shorter, non-linear noise schedules (as few as 6 steps) were successfully used in both works for sampling without observing any significant degradation in generation. However, searching for a highly-performed noise schedule for sampling remains an unsolved issue. Kong et al. \cite{zhifeng2021} defined a fixed, practically useful noise schedule for sampling. As a more general approach, given a well-trained score network Chen et al. \cite{nanxin2020} used a grid search (GS) algorithm to select $\hat{\bbeta}$. Yet, GS is prohibitively slow for $N>6$. Meanwhile, DDPMs \cite{jonathan2019} mentioned that the training noise schedule $\bbeta$ could be learned by reparameterization (although not implemented or investigated in their work). We present an ablation study in Sec. \ref{sec:exp} and demonstrate that a direct reparameterization on $\bbeta$ is less favorable.

In this paper, bilateral denoising diffusion models (BDDMs) are motivated (1) to provide a theoretical grounding on the training performed at the diffusion step $t$, which is inline with the training heuristic in \cite{jonathan2020}, and (2) to allow a direct learning of the noise schedule for sampling based on the theory. The idea underlying BDDMs is to use the forward process to directly sample $\bx_t\sim q_{\bbeta}(\bx_t|\bx_0)$ and then consider the reverse process $p_\theta(\bx_{0:t-1}|\bx_t)$ starting from the sampled $\bx_t$. The joint consideration of forward process and the reverse process in one integral is what we view as the bilateral modeling in BDDMs. Formally, we first define the prior at step $t$ as $\pi(\bx_t)$ such that $p_\theta(\bx_{0:t})=\pi(\bx_t)p_\theta(\bx_{0:t-1}|\bx_{t})$ for $t\in \{2, ..., T\}$. Then, we derive a new lower bound to the log marginal likelihood:
\begin{proposition}
Given a noise schedule $\bbeta$, the following lower bound holds for $t\in \{2, ..., T\}$:
\begin{align}
    \log p_{\theta} (\bx_0)\geq
    \cF_\text{score}^{(t)}(\theta):= -\mathbb{E}_{q_{\bbeta}(\bx_t|\bx_0)} \left[\mathcal{L}^{(t)}_\text{score}(\theta)+\mathcal{R}_\theta(\bx_0, \bx_t)\right],
\end{align}
where 
\begin{align}
    \mathcal{L}^{(t)}_\text{score}(\theta)&:=\mathbb{KL}\left({p_\theta (\bx_{t-1}|\bx_{t})}||{\pi (\bx_{t-1})}\right),\\
    \mathcal{R}_\theta(\bx_0, \bx_t)&:=-\mathbb{E}_{ p_\theta(\bx_1|\bx_t)}\left[\log p_\theta(\bx_{0}|\bx_1)\right].
\end{align}
\end{proposition}
The proof is provided in the Appendix. In practice, $\mathcal{R}_\theta(\bx_0, \bx_t)$ can be approximated by a one-step prediction from $\bx_t$ to $\bx_1$ using the non-Markovian probability density function defined in denoising diffusion implicit models (DDIMs) \cite{jiaming2021}. In this regard, the proposed lower bound $\cF_\text{score}^{(t)}(\theta)$ allows us to consider only one $t$ at each training step for efficient training, which is practically more advantageous than the standard ELBO in Eq. (\ref{eq:elbo}) that entails computing a sum of $T$ KL terms. Below, we show that the proposed bound resembles $\cL_\text{ddpm}^{(t)}(\theta)$ under the following conditions:

\begin{proposition}
\label{prop:theta}
If we set $\pi({\bf x}_{t-1})=q_{\bbeta}({\bf x}_{t-1}|{\bf x}_{t}, {\bf x}_0)$ for $t\in \{2, ..., T\}$, then any optimal solution satisfying $\theta^*=\text{argmin}_\theta \cL^{(t)}_\text{ddpm}(\theta)\, \forall t\in \{1, ..., T\},$ also satisfies $\theta^*=\text{argmax}_\theta\cF^{(t)}_\text{score}(\theta)\, \forall t\in \{2, ..., T\}$.
\end{proposition}
The proof is followed in the Appendix. By this proposition, we see that optimizing $\cL^{(t)}_\text{ddpm}(\theta)$ for training the score network $\theta$ is sufficient to maximize the proposed lower bound $\cF^{(t)}_\text{score}(\theta)$. In this sense, we provide a theoretical grounding for the training heuristic used in DDPMs and DDIMs, which can be seen as optimizing a special case of our derived objective (when setting $\pi({\bf x}_{t-1})=q_{\bbeta}({\bf x}_{t-1}|{\bf x}_{t}, {\bf x}_0)$). In addition, it is practically beneficial that our proposed BDDMs can re-use $\theta$ from any well-trained DDPM or DDIM.

Given that $\theta$ can be trained to maximize the log evidence with the pre-specified noise schedule $\bbeta$, the remaining question of interest in BDDMs is how to find a good enough noise schedule $\hat{\bbeta}\in\real^N$ for sampling for an optimized $\theta^*$. To tackle this problem, we introduce a novel \textit{scheduling network} $\sigma_\phi(\bx_n)$, which is responsible for estimating the ratio between two consecutive noise scales ($\hat{\beta}_{n}$ and $\hat{\beta}_{n+1}$) from the current noisy observation $\bx_n$ for inference in a descending order (from $n=N$ to $1$). To make use of the scheduling, we first need to understand the relationship between the noise scale $\hat{\beta}_{n}$ at step $n$ and the noise scale variables at step $n+1$ (i.e. $\hat{\alpha}_{n+1}$ and $\hat{\beta}_{n+1}$). As a remark, we indeed can derive the range of $\hat{\beta}_{n}$ from its definition: 
\begin{remark}
\label{prop:beta}
Suppose the noise schedule for sampling is monotonic, i.e., $0<\hat{\beta}_{1}<\dotsc < \hat{\beta}_{N}<1$, then, for $1\leq n < N$, $\hat{\beta}_{n}$ satisfies the following inequality:
\begin{align}
     0 < \hat{\beta}_{n} < \min\left\{1 - \frac{\hat{\alpha}_{n+1}^2}{1-\hat{\beta}_{n+1}}, \hat{\beta}_{n+1}\right\}.
\end{align}
\end{remark}
The derivation is presented in the Appendix. Note that monotonic noise schedules have been widely used in the prior arts \cite{yang2019,yang2020improved,yang2021} and would lead to a tighter upper bound for $\hat{\beta}_{n}$. In practice, a tighter bound is helpful to keep the noise schedule short and effective for sampling. Consequently, we define the scheduling network $\sigma_\phi:\real^{D}\mapsto (0,1)$ such that
\begin{align}
\label{eq:stepnet}
    \hat{\beta}_{n} = \min\left\{1 - \frac{\hat{\alpha}_{n+1}^2}{1-\hat{\beta}_{n+1}}, \hat{\beta}_{n+1}\right\} \sigma_\phi(\bx_{n}),
\end{align}
where $\phi$ is a set of learnable neural network parameters. Given the definition of scheduling network, we are now able to estimate the noise schedule for sampling, which we call the \textit{noise scheduling} process. By defining the maximum number of reverse steps (denoted as $M$), we can sequentially compute the score $\beps_\theta\left(\bx_{n}, \hat{\alpha}_{n}\right)$ as in sampling except that the noise scale $\hat{\alpha}_{n}=\frac{\hat{\alpha}_{N}}{\prod_{i=n+1}^{N}\sqrt{1-\hat{\beta}_{i}}}$ is backward computed based on the output of scheduling network starting from the white noise $\bx_N\sim \cN(\zero, \bI)$. The remaining problem is how to learn the network parameter $\phi$. We note that $\cL_\text{score}^{(t)}$ depends on $\beta_t$ but not $\hat{\beta}_n$, thus it cannot simply be used to learn $\phi$. As a result, we first build a link between sampling and training schedules by adding a constraint to the scheduling network:
\begin{align}
\label{eq:tau}
    q_\phi(\bx_{n+1}|\bx_{n}=\tilde{\bx})=q_{\bbeta}(\bx_{t+\tau}|\bx_{t}=\tilde{\bx}),
\end{align}
where $1\leq\tau<T$ is an positive integer, and $\tilde{\bx}$ is an arbitrary diffused variable. This constraint states that one step of diffusion using $\hat{\beta}_n$ estimated by $\sigma_\phi$ equals to $\tau$ steps of diffusion using $\beta_{t}, ..., \beta_{t+\tau}$. Considering the simplest case where $\tau=1$, i.e., $q_\phi(\bx_{n}|\bx_{n-1})=q_{\bbeta}(\bx_{t}|\bx_{t-1})$, we can deduce the following proposition for learning $\phi$:
\begin{proposition}
\label{prop:phi}
Assuming $\theta$ has been optimized and hypothetically converged to the optimal parameters $\theta^*$, where by optimal parameters it means that  $p_{\theta^*}(\bx_{1:t-1}|\bx_0)=q_{\bbeta}(\bx_{1:t-1}|\bx_0)$. Then, we can express the gap between $\log p_{\theta}(\bx_0)$ and $\cF_\text{score}^{(t)}(\theta^{*})$ by $\phi$ as follows:
\begin{align}
\log p_{\theta^{*}}(\bx_0)-\cF_\text{score}^{(t)}(\theta^{*})= \mathbb{E}_{q_{\bbeta}(\bx_{t}|\bx_{0})}\left[\sum_{i=2}^{t}\mathcal{L}_\text{step}^{(i)}(\phi;\theta^{*})\right],
\end{align}
where
\begin{align}
\mathcal{L}_\text{step}^{(i)}(\phi;\theta^{*}):=\mathbb{KL}\left(p_{\theta^*} (\bx_{i-1}|\bx_i) || q_\phi (\bx_{i-1}|\bx_0)\right).
\end{align}
\end{proposition}
The proof is shown in Appendix. From this proposition, we can observe that minimizing $\phi$ over $\cL_\text{step}^{(i)}(\phi;\theta^{*})$ leads to minimizing the gap between $\log p_{\theta^{*}}(\bx_0)$ and $\cF_\text{score}^{(t)}(\theta^{*})$ even when $\theta$ is at its optimal. Besides its practical values, the above proposition is also theoretically appealing. Followed by this proposition, we design a new lower bound $\mathcal{F}^{(t)}_\text{bddm}(\theta, \phi)$ that enjoys nice theoretical properties:
\begin{corollary}
\label{thm:lb}
Relative to the standard ELBO evaluated at step $t\in\{2, ..., T\}$ defined in \cite{jonathan2020} as
\begin{align}
    \mathcal{F}^{(t)}_\text{elbo}(\theta)&:=-\mathbb{E}_{q_{\bbeta}(\bx_t|\bx_0)}\left[\mathbb{KL}\left(q_{\bbeta}(\mathbf{x}_{t-1}|\mathbf{x}_{t}, \mathbf{x}_{0})||p_{\theta}(\mathbf{x}_{t-1}|\mathbf{x}_{t})\right)+\mathcal{R}_\theta(\bx_0, \bx_t)\right],
\end{align}
we propose a new lower bound as the following
\begin{align}
    \mathcal{F}^{(t)}_\text{bddm}(\theta, \phi)&:=\begin{cases}
        \cF_\text{score}^{(t)}(\theta) \quad & \text{if} \quad \theta \neq \theta^{*}\\
        \cF_\text{score}^{(t)}(\theta)+\mathbb{E}_{q_{\bbeta}(\bx_t|\bx_0)}\left[\cL_\text{step}^{(t)}(\phi;\theta)\right] \quad & \text{if} \quad \theta = \theta^{*}
    \end{cases}
\end{align}
which leads to a tighter lower bound when the conditions in Proposition 2 and 3 are satisfied:
\begin{align}
\log p_{\theta}(\mathbf{x}_0)\geq
\mathcal{F}^{(t)}_\text{bddm}(\theta, \phi)\geq \mathcal{F}^{(t)}_\text{elbo}(\theta).
\end{align}
\end{corollary}

The proof can be found in Appendix. Note that using $\mathcal{F}^{(t)}_\text{elbo}(\theta)$ to train the score network by randomly sampling the diffusion step $t$ at each training step has been proven an effective and efficient training heuristic for score-based DPMs, such as DDPM \cite{jonathan2020}, DDIM \cite{jiaming2021} and Improved DDPM \cite{nichol2021improved}. Corollary \ref{thm:lb} suggests that using $\mathcal{F}^{(t)}_\text{bddm}(\theta^{*}, \phi)$ could be a better training method than $\mathcal{F}^{(t)}_\text{elbo}(\theta^{*})$ towards maximizing the log evidence by additionally learning $\phi$ with $\cL_\text{step}^{(t)}(\phi;\theta^*)$. Based on these propositions, we can devise the training and inference algorithms for BDDMs, which are discussed in the following section from a practical point of view.

\section{Training, noise scheduling, and inference}
In BDDMs, we train two neural networks: i) a score network $\beps_\theta$ for sampling, and ii) a scheduling network $\sigma_\phi$ for estimating a noise schedule for sampling. In this section, we first present the training objectives for learning $\theta$ and $\phi$, respectively, based on the above derived propositions.

\subsection{Training objectives}
By Corollary \ref{thm:lb}, we know that $\cF_\text{bddm}^{(t)}$ contains two terms: $\cF_\text{score}^{(t)}(\theta)$ and $\cL_\text{step}^{(t)}(\phi;\theta)$. On one hand, by Proposition \ref{prop:theta}, optimizing $\cF_\text{score}^{(t)}(\theta)$ w.r.t $\theta$ can be simplified into minimizing the loss $\cL_\text{ddpm}^{(t)}(\theta)$ for all $t\in \{1, ..., T\}$, which has been successfully employed in the prior works \cite{jonathan2020}. On the other hand, by Proposition \ref{prop:phi}, we know that $\phi$ should be trained after $\theta$ is well-optimized. In a realistic setting, we can assume the optimized $\theta^*$ to be closed to the hypothetically optimal values. By fixing the well-optimized parameters $\theta^*$, we then optimize the proposed loss $\cL_\text{step}^{(t)}(\phi;\theta^*)$ w.r.t $\phi$, which can be simplified into weighted $\ell_2$ norms:
\begin{align}
    \cL_\text{step}^{(t)}(\phi;\theta^*)&=\frac{1}{2(1-\hat{\beta}_t(\phi)-{\alpha}_{t}^2)}\left\lVert\sqrt{1-\alpha_t^2}\beps_t - \frac{\hat{\beta}_t(\phi)}{\sqrt{1-\alpha_t^2}}\beps_{\theta^*}(\bx_t, \alpha_t) \right\rVert^2_2+C_t(\phi),
\end{align}
where
\begin{align}
    C_t(\phi):=\frac{1}{4}\log \frac{1-{\alpha}^2_t}{\hat{\beta}_t(\phi)}+\frac{D}{2}\left(\frac{\hat{\beta}_t(\phi)}{1-{\alpha}_t^2}-1\right).
\end{align}
The simplification procedures can be found in Appendix.


\algrenewcommand\algorithmicindent{0.5em}%
\begin{figure}[t]
\begin{minipage}[t]{0.49\textwidth}
\setcounter{algorithm}{0}
\begin{algorithm}[H]
  \caption{Training $\theta$} \label{alg:training_theta}
  \small
  \begin{algorithmic}[1]
    \State Given $T, \{\beta_t\}_{t=1}^T$
    \State Compute $\alpha_{t}=\prod_{i=1}^t \sqrt{1-\beta_t}$ from $t=1$ to $T$
    \Repeat
      \State $\bx_0 \sim p_\text{data}(\bx_0)$
      \State $t \sim \mathrm{Uniform}(\{1, \dotsc, T\})$
      \State $\beps_t\sim\mathcal{N}(\zero,\bI)$
      \State $\bx_{t}=\hat{\alpha}_{t} \bx_0 + \sqrt{1-\hat{\alpha}_{t}^2}\beps_t$
      \State $\cL_\text{ddpm}^{(t)}=\left\| \beps_t - \beps_\theta(\bx_{t}, \alpha_{t}) \right\|^2_2$
      \State Take a gradient descent step on $\nabla_\theta \cL_\text{ddpm}^{(t)}$
    \Until{converged}
  \end{algorithmic}
\end{algorithm}
\vspace{-2.54em}
\setcounter{algorithm}{2}
\begin{algorithm}[H]
  \caption{Noise Scheduling} \label{alg:nspred}
  \small
  \begin{algorithmic}[1]
    \State Given $\hat{\alpha}_N, \hat{\beta}_N$
    \State $\bx_N \sim \mathcal{N}(\zero, \bI)$
    \For{$n=N, \dotsc, 2$}
      \State $\bx_{n-1} \sim p_\theta(\bx_{n-1}|\bx_{n};\hat{\alpha}_{n}, \hat{\beta}_{n})$
      \State $\hat{\alpha}_{n-1}=\frac{\hat{\alpha}_{n}}{\sqrt{1-\hat{\beta}_{n}}}$
      \State $\hat{\beta}_{n-1}=\min\{1-\hat{\alpha}_{n-1}^2, \hat{\beta}_{n}\}\sigma_\phi(\bx_{n-1})$
      \If{$\hat{\beta}_{n-1}<\beta_1$}
            \State\textbf{return} $\hat{\beta}_{n}, \dotsc, \hat{\beta}_{N}$
      \EndIf
    \EndFor
    \State \textbf{return} $\hat{\beta}_{1}, \dotsc, \hat{\beta}_{N}$
  \end{algorithmic}
\end{algorithm}
\end{minipage}
\hfill
\begin{minipage}[t]{0.49\textwidth}
\setcounter{algorithm}{1}
\begin{algorithm}[H]
  \caption{Training $\phi$} \label{alg:training_phi}
  \small
  \begin{algorithmic}[1]
    \State Given $\theta^*, \tau, T, \{\alpha_{t}, \beta_t\}_{t=1}^T$
    \Repeat
      \State $\bx_0 \sim p_\text{data}(\bx_0)$
      \State $t \sim \mathrm{Uniform}(\{2, \dotsc, T-\tau\})$
      \State $\hat{\alpha}_{n}={\alpha}_{t},\,\hat{\beta}_{n+1}=1-({\alpha}_{t+\tau}/{\alpha}_{t})^2$
      \State $\delta_{n}=\sqrt{1-\hat{\alpha}_{n}^2}$
      \State $\beps_{n}\sim\mathcal{N}(\zero,\bI)$
      \State $\bx_{n}=\hat{\alpha}_{n} \bx_0 + \delta_{n}\beps_{n}$
      \State $\beps^{(n)}_{\theta^*}=\beps_{\theta^*}(\bx_{n}, \hat{\alpha}_{n})$
      \State $\hat{\beta}_{n}=\min\{\delta_{n}^2, \hat{\beta}_{n+1}\}\sigma_\phi(\bx_{n})$
      \State $C_{n}=4^{-1}\log (\delta_{n}^2/\hat{\beta}_{n})+2^{-1}D\left(\hat{\beta}_{n}/\delta_{n}^2-1\right)$
      \State $\cL_\text{step}^{(n)}=\frac{1}{2(\delta_{n}^2-\hat{\beta}_{n})}\left\lVert \delta_{n}\beps_{n} - \frac{\hat{\beta}_{n}}{\delta_{n}}\beps^{(n)}_{\theta^*}\right\rVert^2_2+C_{n}$
      \State Take a gradient descent step on $\nabla_\phi \cL_\text{step}^{(n)}$
    \Until{converged}
  \end{algorithmic}
\end{algorithm}
\vspace{-2.54em}
\setcounter{algorithm}{3}
\begin{algorithm}[H]
  \caption{Sampling} \label{alg:sampling}
  \small
  \begin{algorithmic}[1]
    \State Given $\{\hat{\beta}_{n}\}_{n=1}^{N}, \bx_{N} \sim \mathcal{N}(\zero, \bI)$
    \State Compute $\alpha_{n}=\prod_{i=1}^n \sqrt{1-\hat{\beta}_{n}}$ from $n=1$ to $N$ 
    \For{$n=N, \dotsc, 1$}
      \State $\bx_{n-1} \sim p_\theta(\bx_{n-1}|\bx_{n};\alpha_{n}, \hat{\beta}_{n})$
    \EndFor
    \State \textbf{return} $\bx_0$
  \end{algorithmic}
\end{algorithm}
\end{minipage}
\vspace{-1.2em}
\end{figure}

\subsection{Efficient training}
Suggested by Proposition 3, we know that $\theta$ should be optimized before learning $\phi$. For training the score network $\beps_\theta$, similar to the prior settings in \cite{jonathan2020, nanxin2020, jiaming2021}, we define $\bbeta$ as a linear noise schedule:
\begin{align}
\label{eq:linsamp}
    \beta_{t}=\frac{\varepsilon}{T-t+1}, \quad \text{for}\quad 1\leq t
    \leq T,
\end{align}
where $\varepsilon$ is a hyperparameter that specifies the maximum value in this linear schedule, i.e., $\beta_T$. By Proposition \ref{prop:theta}, we can sample $t\sim \text{Uniform}(\{1, ..., T\})$ for evaluating $\cL_\text{ddpm}^{(t)}$, resulting in the same Algorithm \ref{alg:training_theta} as in \cite{jonathan2020}.

Next, given a well-optimized $\theta$, we can start training the scheduling network $\sigma_\phi$. By the definition of $\sigma_\phi$ in Eq. (\ref{eq:stepnet}), for training we need to first obtain the noise scale from the preceding step, i.e., $\hat{\beta}_{n+1}$. To achieve this, we randomly sample the noise scale as follows:
\begin{align}
    \hat{\beta}_{n+1} \sim \text{Uniform}\left(\left\{  1-({\alpha}_{t+\tau}/{\alpha}_{t})^2\,|\,t\in\{2, ..., T-\tau\}\right\}\right),
\end{align}
where $0<\tau<T$ is a hyperparameter defined in Eq. (\ref{eq:tau}) to control the inference step size relative to the noise schedule for training $\bbeta$. Expectedly, a larger $\tau$ should lead to a shorter noise schedule $\hat{\bbeta}$. Here, we note that Corollary 1 still holds even if we change the index from $t$ to $n$. Followed from the this sampling strategy, we can efficiently train the scheduling network, as presented in Algorithm \ref{alg:training_phi}. Our empirical observation in section \ref{sec:exp} shows that, although a linear schedule is used to define $\bbeta$ to sample the local transition from $n+1$ to $n$, the predicted schedule $\hat{\bbeta}$ is not limited into a linear one.

\subsection{Noise scheduling for fast and high-quality sampling}
After the score network and the scheduling network are trained, BDDMs split the inference procedure into two phases: (1) the noise scheduling phase and (2) the sampling phase. First, we consider sampling from a reverse process with $N$ iterations maximum to obtain a noise schedule for sampling $\hat{\bbeta}$. Different from the forward computed $\alpha_t$ which is fixed to start from $\sqrt{1-\beta_1}$, $\hat{\alpha}_n$ is a backward computed variable (from $N$ to $1$) that may deviates from the forward one because $\{\hat{\beta}_i\}_i^{n-1}$ are unknown in the noise scheduling phase. Therefore, for noise scheduling, we have two hyperparameters -- $(\hat{\alpha}_N, \hat{\beta}_N)$. Overall, the noise scheduling procedure is presented in Algorithm \ref{alg:nspred}. Note that we use $\beta_1$, the smallest noise scale seen in training, as a threshold to determine when to stop the noise scheduling process, so that we can ignore the small noise scales ($< \beta_1$) unseen by the score network that may lead to numerical issue. By doing so, we can also get a reasonably shorter $\hat{\bbeta}$.
\par
In practice, we apply a grid search algorithm of $M$ bins to find a good set of initial values for $(\hat{\alpha}_N, \hat{\beta}_N)$, which takes $\mathcal{O}(M^2)$ time. This is computationally feasible in contrast to the unscalable $\mathcal{O}(M^N)$ grid search algorithm used in \cite{nanxin2020} that searches over entire noise schedule $\hat{\bbeta}$. Empirically, the grid search for our noise scheduling algorithm can be evaluated on a small subset of the training samples (as few as $1$ sample) to find a generally well-performed noise schedule. 
\par
After the noise scheduling procedure is finished, we fix the noise schedule $\hat{\bbeta}\in\real^N$ to the best searched result for the subsequent sampling over the whole testing set. The sampling procedure is the same as in \cite{jonathan2020}. By initializing the white noise $\bx_N\sim\cN(\zero, \bI)$, we can generate $\bx_0$ after $N$ iterations of the reverse process, as shown in Algorithm \ref{alg:sampling}.

\section{Experiments}
\label{sec:exp}
A series of experiments on generative modeling have been conducted to evaluate BDDMs and to compare it against other recent SOTA denoising diffusion models, including the DDPMs \cite{jonathan2020} and the DDIMs \cite{jiaming2021}. In addition, we used an even more recently proposed noise estimation (NE) approach \cite{san2021noise} that improved the noise scheduling in DDPMs and DDIMs as another strong baseline. The NE approach trains a noise estimator $\hat{\alpha}_t=g(\bx_t)$ to directly predict $\alpha_t^2$ by using a log-scale regression loss $\cL_\text{ne}=\left\| \log(1-\alpha_t^2)-\log(1-\hat{\alpha}_t^2)\right\|_2^2$. At inference time, NE requires a pre-defined noise schedule, e.g., a linear schedule or a Fibonacci schedule.


We evaluated the models on two generative tasks -- speech synthesis and image generation. For speech synthesis, we used the benchmark LJ speech \cite{keith2017} and the VCTK \cite{yamagishi2019vctk} datasets. For image generation, we employed the benchmark CIFAR-10 (32$\times$32) \cite{Krizhevsky09learningmultiple} and CelebA (64$\times$64) datasets. In this section, we present the experimental results on speech synthesis using the LJ speech dataset for analyzing the model behavior of BDDMs and for comparing BDDMs against the prior arts. All models were trained on the same LJ speech training set as in \cite{nanxin2020}. Note that the evaluation metrics in image generation normally entails calculating the statistics over a set of generated images, whereas in speech synthesis we can score each generated audio alone with objective and subjective metrics, which is more convenient for analyzing the model performance.

\begin{table}[t]
\centering
\caption{Performances of different noise schedules on the single-speaker LJ speech dataset, each of which used the same score network \cite{nanxin2020} $\beps_{\theta}(\cdot)$ that was trained for about 1M iterations.
}
\label{tab:all}
\begin{tabular}{lccccc}
 \toprule
        {\bfseries Noise schedule} & \bfseries LS-MSE ($\downarrow$) & \bfseries MCD ($\downarrow$) &\bfseries STOI ($\uparrow$) &\bfseries PESQ ($\uparrow$) & \bfseries MOS ($\uparrow$) \\
 \midrule
 {\bf DDPM \cite{jonathan2020, nanxin2020}} \\
 \quad 8 steps (Grid Search) & 99.8 & 2.33 &  0.938 & 3.21 & 4.25 $\pm$ 0.06\\
 \quad 1,000 steps (Linear) & 81.2 & 2.02 & 0.948 & 3.29 & 4.39 $\pm$ 0.05\\
 \midrule
 \multicolumn{6}{l}{\bf DDIM \cite{jiaming2021}} \\
 \quad 8 steps (Linear) & 128 & 2.63 & 0.930 & 2.93 & 4.15 $\pm$ 0.05\\
 \quad 16 steps (Linear) & 114.6 & 2.49 & 0.942 & 3.06 & 4.30 $\pm$ 0.06\\
 \quad 21 steps (Linear) & 114.2 & 2.49 & 0.945 & 3.10 & 4.31 $\pm$ 0.04\\
 \quad 100 steps (Linear) & 98.8 & 2.32 & \bf 0.954 & 3.27 & 4.44 $\pm$ 0.03\\
 \midrule
 \multicolumn{6}{l}{\bf NE \cite{san2021noise}} \\
 \quad 8 steps (Linear) & 141 & 2.78 & \bf 0.940 & 3.12 & 4.05 $\pm$ 0.06 \\ 
 \quad 16 steps (Linear) & 77.8 & 1.99 & 0.947 & 3.28 & 4.29 $\pm$ 0.06\\
 \quad 21 steps (Linear) & 188 & 3.10 & 0.937 & 2.86 & 3.91 $\pm$ 0.05\\
 \midrule
 \multicolumn{6}{l}{\bf BDDM $(\hat{\alpha}_N,\hat{\beta}_N)$} \\
 \quad 8 steps $(0.3, 0.9)$ & \bf 91.3 & \bf 2.19 & 0.936 & \bf 3.22 & \bf 4.27 $\pm$ 0.04\\
 \quad 16 steps $(0.7, 0.1)$ & \bf 73.3 & \bf 1.88 & \bf 0.949 & \bf 3.32 & \bf 4.36 $\pm$ 0.05\\
 \quad 21 steps $(0.5, 0.1)$ & \bf 72.2 & \bf 1.91 & \bf 0.950 & \bf 3.33 & \bf 4.47 $\pm$ 0.04 \\
 \midrule
 \multicolumn{6}{l}{\bf{Ablated BDDM} (Directly learning $\hat{\bbeta}$)}  \\
 \quad 8 steps & 132 & 2.26 & 0.924 & 3.01 & 4.11 $\pm$ 0.05\\
 \bottomrule
\end{tabular}
\vspace{-1.2em}
\end{table}


For the model architecture, we used the same architecture as in \cite{nanxin2020} for the score network;
we adopted a lightweight GALR network \cite{lam2021effective} for the scheduling network. GALR was originally proposed for speech enhancement, so we considered it well suited for predicting the noise scales. For the configuration of the GALR network, we used a window length of 8 samples for encoding, a segment size of 64 for segmentation and only two GALR blocks of 128 hidden dimensions, and other settings were inherited from \cite{lam2021effective}. To make the scheduling network output with a proper range and dimension, we applied a sigmoid function to the last block's output of the GALR network. Then the result was averaged over the segments and the feature dimensions to obtain the predicted ratio: $\sigma_\phi(\bx)=\text{AvgPool2D}(\sigma(\text{GALR}(\bx)))$, where $\text{GALR}(\cdot)$ denotes the GALR network, $\text{AvgPool2D}(\cdot)$ denotes the average pooling operation applied to the segments and the feature dimensions, and $\sigma(x):=1/(1+e^{-x})$. The same network architecture was used for the NE approach for estimating $\alpha_t^2$ and was shown comparatively better than the ConvTASNet used in the original paper \cite{san2021noise}. It is also notable that the computational cost of a scheduling network is indeed fractional compared to the cost of a score network, as predicting a noise scalar variable is intrinsically a relatively much easier task. Our GALR-based scheduling network, while being able to produce stable and reliable results, was about 3.6 times faster than the score network, meaning that training BDDMs can be almost as fast as training DDPMs or DDIMs. More details regarding the model architecture, the total amount of computing, and the type of resources used can be found in supplementary materials.

\subsection{Sampling quality in objective and subjective metrics}
Since the objective and subjective metrics have been widely applied and well-established for speech quality evaluation, we consider it convincing by evaluating generative models on speech samples. 
Specifically, we used the objective metrics in \cite{nanxin2020} -- the log-Mel spectrogram mean squared error (LS-MSE) and the Mel-cepstral distance (MCD) \cite{kubichek1993mel} to assess the consistency between the original waveform and the generated waveform in the Mel-frequency domain. In addition, to measure the noisiness and the distortion of the generated speech relative to the reference speech, we adopted two commonly used metrics in speech enhancement -- the perceptual evaluation of speech quality (PESQ) \cite{rix2001perceptual} and the short-time objective intelligibility (STOI) measure \cite{taal2010short}. Mean opinion scores (MOS) were used as the subjective metric on the speech quality.


The result is present in Table \ref{tab:all}, where we include the performance of three sets of $(\alpha_N, \beta_N)$ corresponding to 8, 16, and 21 inference steps. Remarkably, our proposed BDDMs outperformed the 1,000-step DDPM with only 16 or 21 steps, which is conceivably tolerable for a fast and high-quality generation. Noticeably, the DDIM with 100 steps (10x acceleration) showed superior generative performance than the DDPM with 1000 steps. Yet, while its inference was accelerated by taking only 8, 16, or 21 inference steps, the corresponding performances failed to compete with our BDDM. Although the performance of the NE approach seems promising with 16 steps, it surprisingly degraded drastically with 21 steps. In comparison, the performance of BDDM was found to be much stabler than the NE approach. Since the same score network was applied to all the methods, we can confirm that noise scheduling is vital for improving the sampling quality and the efficiency of denoising diffusion models.



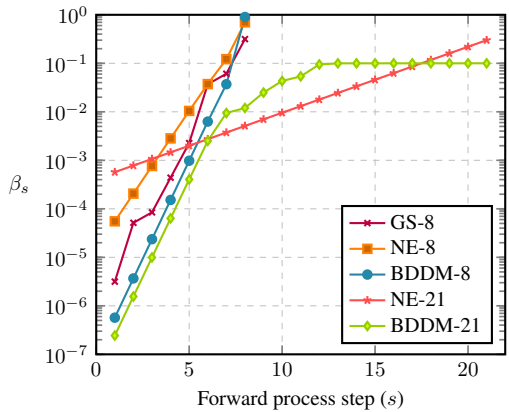
\begin{figure}[t]
\begin{minipage}[H]{0.49\textwidth}
\vspace{-0.1em}
\begin{figure}[H]
    \centering
\resizebox{\columnwidth}{!}
{
    \begin{tikzpicture}
    \begin{axis}[
        line width=1pt,
        xlabel={Forward process step ($s$)},
        ylabel={$\beta_s$},
        ylabel style={rotate=-90},
        xmin=0, xmax=22,
        ymin=1e-7, ymax=1.,
        ymode=log,
        legend pos=south east,
        ymajorgrids=true,
        xmajorgrids=true,
        grid style=dashed,
        legend cell align={left},
        line width=1pt,
        tick style={line width=0.8pt},
        cycle list name=exotic,
    ]
    
    \addplot
    [
        color=purple,
        mark=x,
        mark options={solid},
        ]
        coordinates {
            ( 1 , 3.1524424717774302e-06 )
            ( 2 , 5.1471661894049195e-05 )
            ( 3 , 8.457044163898543e-05 )
            ( 4 , 0.00043803046108052555 )
            ( 5 , 0.0022687676818985555 )
            ( 6 , 0.03704341763070655 )
            ( 7 , 0.06086413520695782 )
            ( 8 , 0.315244247177743 )
        };
    \addlegendentry{GS-8}
    \addplot
        coordinates {
            ( 1 , 5.52524507e-05 )
            ( 2 , 2.05671589e-04 )
            ( 3 , 7.64078635e-04 )
            ( 4 , 2.83316965e-03 )
            ( 5 , 1.04246298e-02 )
            ( 6 , 3.73152755e-02 )
            ( 7 , 1.21843159e-01 )
            ( 8 , 6.99999988e-01 )
        };
    \addlegendentry{NE-8}
    \addplot
        coordinates {
            (1,5.696867333426781e-07)
            (2,3.6706858281831956e-06)
            (3,2.3651098672416992e-05)
            (4,0.00015237453044392169)
            (5,0.0009810632327571511)
            (6,0.006289184559136629)
            (7,0.03708929196000099) 
            (8,0.9)
        };
    \addlegendentry{BDDM-8}
    \addplot
        coordinates {
        ( 1 , 0.000570381002034992 )
        ( 2 , 0.0007802124600857496 )
        ( 3 , 0.0010672365315258503 )
        ( 4 , 0.0014598509296774864 )
        ( 5 , 0.0019968999549746513 )
        ( 6 , 0.002731518354266882 )
        ( 7 , 0.0037363877054303885 )
        ( 8 , 0.0051109278574585915 )
        ( 9 , 0.006991133093833923 )
        ( 10 , 0.009563026949763298 )
        ( 11 , 0.01308106817305088 )
        ( 12 , 0.017893323674798012 )
        ( 13 , 0.024475907906889915 )
        ( 14 , 0.03348008915781975 )
        ( 15 , 0.045796722173690796 )
        ( 16 , 0.06264439225196838 )
        ( 17 , 0.085689976811409 )
        ( 18 , 0.11721355468034744 )
        ( 19 , 0.1603340059518814 )
        ( 20 , 0.21931758522987366 )
        ( 21 , 0.30000001192092896 )
        };
    \addlegendentry{NE-21}
    \addplot
        coordinates {
            ( 1 , 2.438912929392245e-07 )
            ( 2 , 1.5570896039207582e-06 )
            ( 3 , 9.940195013768971e-06 )
            ( 4 , 6.34239986538887e-05 )
            ( 5 , 0.00040336253005079925 )
            ( 6 , 0.002510148799046874 )
            ( 7 , 0.009438818320631981 )
            ( 8 , 0.011978118680417538 )
            ( 9 , 0.024927008897066116 )
            ( 10 , 0.043224889785051346 )
            ( 11 , 0.05376734584569931 )
            ( 12 , 0.09471277892589569 )
            ( 13 , 0.09984925389289856 )
            ( 14 , 0.099870465695858 )
            ( 15 , 0.0998896136879921 )
            ( 16 , 0.09990809857845306 )
            ( 17 , 0.09992634505033493 )
            ( 18 , 0.09994460642337799 )
            ( 19 , 0.09996292740106583 )
            ( 20 , 0.09998137503862381 )
            ( 21 , 0.10000000149011612 )
        };
    \addlegendentry{BDDM-21}
    \end{axis}
    \end{tikzpicture}
    }
    \caption{Noise schedules in log-scale y axis}
    \label{fig:ns}
    \end{figure}
\end{minipage}
\begin{minipage}[H]{0.493\textwidth}
\begin{figure}[H]
    \centering
\resizebox{\columnwidth}{!}
{
    \begin{tikzpicture}
    \begin{axis}[
        xlabel={Reverse process step ($S-s$)},
        ylabel={Distortion (PESQ)},
        line width=1pt,
        tick style={line width=0.8pt},
        xmin=0, xmax=35,
        ymin=0, ymax=3.5,
        legend pos=south east,
        ymajorgrids=true,
        xmajorgrids=true,
        grid style=dashed,
        legend cell align={left},
        cycle list name=exotic,
    ]
    
    \addplot
    [
        color=purple,
        mark=x,
        mark options={solid},
        ]
        coordinates {
            ( 1 , 0.8294792771339417 )
            ( 2 , 0.8530141115188599 ) 
            ( 3 , 1.4403579235076904 ) 
            ( 4 , 1.9269015789031982 ) 
            ( 5 , 2.3134193420410156 ) 
            ( 6 , 2.5593831539154053 ) 
            ( 7 , 3.039738416671753 ) 
            ( 8 , 3.251433849334717 ) 
        };
    \addlegendentry{GS-8}
    \addplot
        coordinates {
            ( 1 , 0.5919641256332397 )
            ( 2 , 0.8093312382698059 ) 
            ( 3 , 1.0919444561004639 ) 
            ( 4 , 1.4105643033981323 ) 
            ( 5 , 1.8015111684799194 ) 
            ( 6 , 2.1903324127197266 ) 
            ( 7 , 2.59669828414917 ) 
            ( 8 , 3.2087786197662354 ) 
        };
    \addlegendentry{NE-8}
    \addplot
        coordinates {
            ( 1 , 0.5116003155708313 )
            ( 2 , 0.6860295534133911 ) 
            ( 3 , 0.744994580745697 ) 
            ( 4 , 0.9119074940681458 ) 
            ( 5 , 1.0512703657150269 ) 
            ( 6 , 1.2143681049346924 ) 
            ( 7 , 1.4608231782913208 ) 
            ( 8 , 2.7988202571868896 ) 
        };
    \addlegendentry{DDIM-8}
    \addplot
        coordinates {
            ( 1 , 0.6467728018760681 )
            ( 2 , 0.9569783210754395 )
            ( 3 , 1.5158710479736328 )
            ( 4 , 2.0660274028778076 )
            ( 5 , 2.548900842666626 )
            ( 6 , 2.953305959701538 )
            ( 7 , 3.2330310344696045 )
            ( 8 , 3.339324951171875 )
        };
    \addlegendentry{BDDM-8}
    \addplot
        coordinates {
        ( 1 , 0.9096967577934265 )
        ( 2 , 0.6921365857124329 )
        ( 3 , 0.7677467465400696 )
        ( 4 , 0.7952574491500854 )
        ( 5 , 0.8376625776290894 )
        ( 6 , 0.8531847596168518 )
        ( 7 , 0.908385694026947 )
        ( 8 , 0.9238301515579224 )
        ( 9 , 0.9701539874076843 )
        ( 10 , 0.9975759387016296 )
        ( 11 , 1.067462682723999 )
        ( 12 , 1.1199296712875366 )
        ( 13 , 1.1724032163619995 )
        ( 14 , 1.2115734815597534 )
        ( 15 , 1.297023057937622 )
        ( 16 , 1.3945307731628418 )
        ( 17 , 1.535074234008789 )
        ( 18 , 1.7255051136016846 )
        ( 19 , 2.0400478839874268 )
        ( 20 , 3.0073440074920654 )
        ( 21 , 3.1349079608917236 )
        };
    \addlegendentry{DDIM-21}
    \addplot
        coordinates {
        ( 1 , 1.5139533281326294 )
        ( 2 , 0.7797483801841736 )
        ( 3 , 0.43406543135643005 )
        ( 4 , 0.48081275820732117 )
        ( 5 , 1.0297622680664062 )
        ( 6 , 1.0572447776794434 )
        ( 7 , 0.7007320523262024 )
        ( 8 , 0.7391523122787476 )
        ( 9 , 0.8205870985984802 )
        ( 10 , 0.8745479583740234 )
        ( 11 , 0.9477370977401733 )
        ( 12 , 1.0190973281860352 )
        ( 13 , 1.0830949544906616 )
        ( 14 , 1.1730924844741821 )
        ( 15 , 1.2523056268692017 )
        ( 16 , 1.3599696159362793 )
        ( 17 , 1.4704370498657227 )
        ( 18 , 1.6053075790405273 )
        ( 19 , 1.7887353897094727 )
        ( 20 , 2.046849012374878 )
        ( 21 , 2.998659133911133 )
        };
    \addlegendentry{NE-21}
    \addplot
        coordinates {
            ( 1 , 1.302849292755127 )
            ( 2 , 1.0607725381851196 )
            ( 3 , 1.1455035209655762 )
            ( 4 , 1.2608511447906494 )
            ( 5 , 0.6129055023193359 )
            ( 6 , 0.5531123876571655 )
            ( 7 , 0.810085117816925 )
            ( 8 , 0.8839606642723083 )
            ( 9 , 0.5743933916091919 )
            ( 10 , 0.6749069690704346 )
            ( 11 , 0.7473670244216919 )
            ( 12 , 0.8522590398788452 )
            ( 13 , 1.0300406217575073 )
            ( 14 , 1.180311679840088 )
            ( 15 , 1.5316863059997559 )
            ( 16 , 2.054168701171875 )
            ( 17 , 2.5304880142211914 )
            ( 18 , 2.9250988960266113 )
            ( 19 , 3.2110636234283447 )
            ( 20 , 3.338289499282837 )
            ( 21 , 3.3549654483795166 )
        };
    \addlegendentry{BDDM-21}
    \end{axis}
    \end{tikzpicture}
    }
    \caption{PESQ scores of different methods}
    \label{fig:pesq}
    \end{figure}
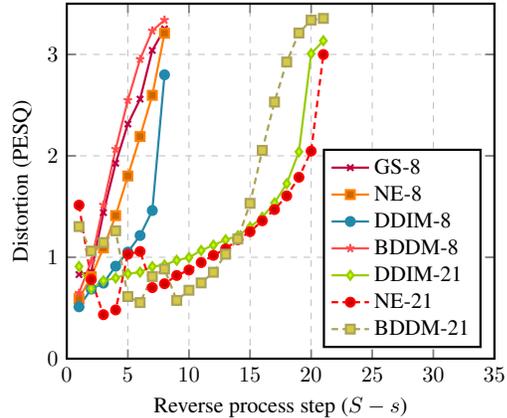
\end{minipage}
\vspace{-1.2em}
\end{figure}

\subsection{Analyzing noise scheduling behaviors}
By looking into BDDMs' behavior of noise scheduling, we want to provide more insights about this work's core novelty and the critical component for achieving the promising performance. In this sense, we compared the noise schedules generated by different methods in a log-scale plot, as illustrated in Figure \ref{fig:ns}, in which the ``GS'' is referred to as the grid search method for DDPM, and the number after the model name is referred to as steps of the model presented in Table \ref{tab:all}. We ask the readers to relate the performance of each model in Table \ref{tab:all} to the plot of its actual noise schedule. We can observe that the ``NE-21'' had the ``flattest'' schedule, which may explain why it also had the lowest scores in both objective and subjective metrics. In contrast, the best performing ``BDDM-21'' estimated the steepest schedule and, interestingly, devised a turning point after the noise scale went greater than about 0.01. We also observed a less obvious turning behavior from the noise schedules of ``BDDM-8'' after passing 0.1. Before the turning point, our BDDMs tended to keep the noise scales as a geometric sequence. We conceive that such behavior is impressive as the step-index was unknown to the scheduling network $\sigma_\phi$. In Figure \ref{fig:pesq}, we also compared the change of PESQ as a reliable measure of distortion, as the sampling index $s$ started from $S$ to $1$ on the same held-out example among different methods. Besides having the highest final PESQ score, BDDMs quickly surpassed all other approaches after about a half of the total sampling steps.

\subsection{Ablation studies}
We attribute the primary superiority of BDDMs to the proposition of the scheduling network for modeling the local transition of noise scales and the loss function for learning $\phi$. In this section, we attempt to study an ablation of these propositions. First, we ablated the scheduling network by setting $\bbeta$ as a learnable parameter, such that our proposed loss function for learning the steps became $\cL_\text{step}^{(t)}(\hat{\bbeta};\theta)$. We reported the corresponding performance in the last row of Table \ref{tab:all}, which appeared to be worse than the scheduling-network-based generations by comparing the case of 8 steps. Besides, we also noted that directly learning $\hat{\bbeta}$ was not scalable to large steps due to the memory constraint for back-propagation. Furthermore, we also tried to keep the scheduling network using alternative training losses, including the negative ELBO $-\cF_\text{elbo}^{(t)}$. However, it turned out the network would degenerate in these cases that it predicted an all-one noise schedule. 

\section{Conclusions}
In this paper, we introduced the bilateral denoising diffusion models (BDDMs), a novel generative model that simultaneously parameterized the forward process and the reverse process by two specific networks: the score network $\beps_\theta$ and the scheduling network $\sigma_\phi$, respectively. We derived a new lower bound to the log evidence for learning $\theta$ that can be efficiently computed for one time step $t$ rather than processing though $T$ time steps. We also showed that the new objective could lead to the same loss derived in DDPMs \cite{jonathan2020} under a reasonable condition. We then derived another training objective for learning $\phi$, which can be viewed as tightening the lower bound when $\theta$ is optimized. Based on the new objectives, we designed a new lower bound tighter than the standard evidence lower bound. Moreover, an efficient training algorithm and a noise scheduling algorithm were respectively presented based on the propositions. Finally, we demonstrated the superiority of BDDMs over previous denoising diffusion models in terms of both generation quality and sampling speed.

\bibliographystyle{IEEEtran}
\bibliography{mybib}


\newpage
\appendix
\section{Mathematical proofs for deriving BDDMs}

\begin{proposition}
Given a noise schedule $\bbeta$, the following lower bound holds for $t\in \{2, ..., T\}$:
\begin{align}
    \log p_{\theta} (\bx_0)\geq
    \cF_\text{score}^{(t)}(\theta):= -\mathbb{E}_{q_{\bbeta}(\bx_t|\bx_0)} \left[\mathcal{L}^{(t)}_\text{score}(\theta)+\mathcal{R}_\theta(\bx_0, \bx_t)\right],
\end{align}
where 
\begin{align}
    \mathcal{L}^{(t)}_\text{score}(\theta)&:=\mathbb{KL}\left({p_\theta (\bx_{t-1}|\bx_{t})}||{\pi (\bx_{t-1})}\right),\\
    \mathcal{R}_\theta(\bx_0, \bx_t)&:=-\mathbb{E}_{ p_\theta(\bx_1|\bx_t)}\left[\log p_\theta(\bx_{0}|\bx_1)\right].
\end{align}
\end{proposition}
\begin{proof}
\begin{align}
\log p_{\theta}(\bx_{0})=&\log\int p_\theta(\bx_{0:t-1}) d\bx_{1:t-1}\\
=&\log\int p_\theta(\bx_{0:t-1})\left(\int \frac{p_\theta(\bx_{1:t-1}|\bx_t)}{p_\theta(\bx_{1:t-1}|\bx_t)} q_{\bbeta}(\bx_t|\bx_0) d \bx_t\right) d\bx_{1:t-1}\\
=&\log\mathbb{E}_{q_{\bbeta}(\bx_t|\bx_0)}\mathbb{E}_{p_\theta(\bx_{1:t-1}|\bx_t)}\left[\frac{p_\theta(\bx_{0:t-1})}{p_\theta(\bx_{1:t-1}|\bx_t)}\right]\\
=&\log\mathbb{E}_{q_{\bbeta}(\bx_t|\bx_0)}\mathbb{E}_{p_\theta(\bx_{1,t-1}|\bx_t)}\left[\frac{p_\theta(\bx_{0}|\bx_1)\pi (\bx_{t-1})}{p_\theta(\bx_{t-1}|\bx_t)}\right]\\
\text{[Jensen's Inequality]}\,\,\geq &\mathbb{E}_{q_{\bbeta}(\bx_t|\bx_0)}\mathbb{E}_{ p_\theta(\bx_1, \bx_{t-1}|\bx_t)}\left[\log\frac{p_\theta(\bx_{0}|\bx_1)\pi (\bx_{t-1})}{p_\theta(\bx_{t-1}|\bx_t)}\right]\\ =&\mathbb{E}_{q_{\bbeta}(\bx_t|\bx_0)}\left[\mathbb{E}_{ p_\theta(\bx_1|\bx_t)}\left[\log p_\theta(\bx_{0}|\bx_1)\right]-\mathbb{KL}\left({p_\theta (\bx_{t-1}|\bx_{t})}||{\pi (\bx_{t-1})}\right)\right]\\
=&-\mathbb{E}_{q_{\bbeta}(\bx_t|\bx_0)}\left[\mathcal{R}_\theta(\bx_0, \bx_t)+\cL^{(t)}_\text{score}(\theta)\right]
\end{align}
\end{proof}

\begin{proposition}
If we set $\pi({\bf x}_{t-1})=q_{\bbeta}({\bf x}_{t-1}|{\bf x}_{t}, {\bf x}_0)$ for $t\in \{2, ..., T\}$, then any optimal solution satisfying $\theta^*=\text{argmin}_\theta \cL^{(t)}_\text{ddpm}(\theta)\, \forall t\in \{1, ..., T\},$ also satisfies $\theta^*=\text{argmax}_\theta\cF^{(t)}_\text{score}(\theta)\, \forall t\in \{2, ..., T\}$.
\end{proposition}
\begin{proof}
By definition, we have
\begin{align}
\label{eq:gaussian1}
    p_\theta(\bx_{t-1}|\bx_{t})
    =
    &\cN
    \left(
        \frac
        {1}{\sqrt{1-\beta_{t}}}
        \left(
        \bx_{t}
        -
        \frac
        {\beta_{t}}
        {\sqrt{1-\alpha_{t}^2}}
        \beps_\theta\left(\bx_{t}, \alpha_{t}\right)
        \right),
        \frac
        {(1-\alpha_{t-1})}
        {(1-\alpha_{t})}\beta_{t}
        \bI
    \right),\\
\label{eq:gaussian2}
    q_{\bbeta}(\bx_{t-1}|\bx_{t}, \bx_{0})
    =
    &\cN
    \left(
        \frac
        {\alpha_{t-1}\beta_t}{1-\alpha_t^2}\bx_{0}
        +
        \frac
        {\sqrt{1-\beta_t}(1-\alpha_{t-1}^2)}{1-\alpha_t^2}\bx_{t}
        ,
        \frac
        {(1-\alpha_{t-1})}
        {(1-\alpha_{t})}\beta_{t}
        \bI
    \right)\\
    =
    &\cN
    \left(
        \frac
        {\alpha_{t-1}\beta_t}{1-\alpha_t^2}
        \frac
        {\bx_t-\sqrt{1-\alpha_t^2}\beps_t}{\alpha_t}
        +
        \frac
        {\sqrt{1-\beta_t}(1-\alpha_{t-1}^2)}{1-\alpha_t^2}\bx_{t}
        ,
        \sigma_{t}^2
        \bI
    \right)\\
    =
    &\cN
    \left(
        \left(
        \frac
        {\alpha_{t-1}\beta_t}{\alpha_{t}(1-\alpha_t^2)}
        +
        \frac
        {\sqrt{1-\beta_t}(1-\alpha_{t-1}^2)}{1-\alpha_t^2}
        \right)
        \bx_t
        -
        \frac
        {\alpha_{t-1}\beta_t}{\alpha_{t}\sqrt{1-\alpha_t^2}}\beps_t
        ,
        \sigma_{t}^2
        \bI
    \right)\\
    =
    &\cN
    \left(
        \frac
        {\alpha_{t-1}\beta_t+\alpha_{t}\sqrt{1-\beta_t}(1-\alpha_{t-1}^2)}
        {\alpha_{t}(1-\alpha_t^2)}
        \bx_t
        -
        \frac
        {\alpha_{t-1}\beta_t}{\alpha_{t}\sqrt{1-\alpha_t^2}}\beps_t
        ,
        \sigma_{t}^2
        \bI
    \right).
\end{align}
Given that $\pi({\bf x}_{t-1}):=q_{\bbeta}({\bf x}_{t-1}|{\bf x}_{t}, {\bf x}_0)$, we now can simplify $\cL^{(t)}_\text{score}(\theta)$ and $\mathcal{R}_\theta(\bx_0, \bx_t)$ as follows.

Firstly, to express $\cL^{(t)}_\text{score}(\theta)$, we know that $p_\theta (\bx_{t-1}|\bx_{t})$ and $q_{\bbeta}(\bx_{t-1}|\bx_{t}, \bx_{0})$ are two isotropic Gaussians with the same variance, thus the KL becomes a scaled squared error between their means:
\begin{align}
\cL^{(t)}_\text{score}(\theta):=&\KL
    \left(
        p_\theta (\bx_{t-1} |\bx_{t})
        ||
        q_{\bbeta}(\bx_{t-1}|\bx_{t}, \bx_{0})
    \right)\\
    =&
    \frac
    {(1-\beta_t)(1-\alpha_{t}^2)}
    {2(1-\beta_t-\alpha_{t}^2)\beta_{t}}
    \left\lVert 
        \frac
        {\alpha_{t-1}\beta_t+\alpha_{t}\sqrt{1-\beta_t}(1-\alpha_{t-1}^2)}
        {\alpha_{t}(1-\alpha_t^2)}
        \bx_t
        -
        \frac
        {\alpha_{t-1}\beta_t}{\alpha_{t}\sqrt{1-\alpha_t^2}}\beps_t\right.\\
    &\qquad\qquad\qquad\qquad
    \left.-
        \frac
        {1}{\sqrt{1-\beta_{t}}}
        \left(
        \bx_{t}
        -
        \frac
        {\beta_{t}}
        {\sqrt{1-\alpha_{t}^2}}
        \beps_\theta\left(\bx_{t}, \alpha_{t}\right)
        \right)
    \right\rVert_2^2\\
    =&
    \frac
    {(1-\beta_t)(1-\alpha_{t}^2)}
    {2(1-\beta_t-\alpha_{t}^2)\beta_{t}}
    \left\lVert 
    \left(
        \frac
        {\alpha_{t-1}\beta_t+\alpha_{t}\sqrt{1-\beta_t}(1-\alpha_{t-1}^2)}
        {\alpha_{t}(1-\alpha_t^2)}
        -
        \frac
        {1}{\sqrt{1-\beta_{t}}}
    \right)
        \bx_t
        \right.\\
    &\qquad\qquad\qquad\qquad
    \left.-
        \frac
        {\alpha_{t-1}\beta_t}{\alpha_{t}\sqrt{1-\alpha_t^2}}\beps_t
        +
        \frac
        {\beta_{t}\beps_\theta\left(\bx_{t}, \alpha_{t}\right)}
        {\sqrt{(1-\alpha_{t}^2)(1-\beta_{t})}}
    \right\rVert_2^2,\\
    =&
    \frac
    {(1-\beta_t)(1-\alpha_{t}^2)}
    {2(1-\beta_t-\alpha_{t}^2)\beta_{t}}
    \left\lVert 
        0\cdot
        \bx_t
        -
        \frac
        {\beta_{t}\beps_t}
        {\sqrt{(1-\alpha_{t}^2)(1-\beta_{t})}}
        +
        \frac
        {\beta_{t}\beps_\theta\left(\bx_{t}, \alpha_{t}\right)}
        {\sqrt{(1-\alpha_{t}^2)(1-\beta_{t})}}
    \right\rVert_2^2\\
    =&
    \frac
    {(1-\beta_t)(1-\alpha_{t}^2)}
    {2(1-\beta_t-\alpha_{t}^2)\beta_{t}}
    \frac
    {\beta_{t}^2}
    {(1-\alpha_{t}^2)(1-\beta_{t})}
    \left\lVert \beps_t-\beps_\theta\left(\bx_{t}, \alpha_{t}\right)\right\rVert_2^2\\
    =&
    \frac{\beta_{t}}{2(1-\beta_{t}-\alpha_{t}^2)}\left\lVert \beps_t-\beps_\theta\left(\bx_{t}, \alpha_{t}\right)\right\rVert_2^2,
\end{align}
which is proportional to $\cL^{(t)}_\text{ddpm}:=\| \beps_{t} - \beps_\theta\left(\bx_{t}, \alpha_{t}\right)\|^2_2$.

Next, we show that $\mathcal{R}_\theta(\bx_0, \bx_t)$ can be expressed in terms of $\cL^{(t)}_\text{ddpm}$ evaluated at $t=1$:
\begin{align}
    \mathcal{R}_\theta(\bx_0, \bx_t):=&-\mathbb{E}_{p_\theta({\bf x}_{1}|{\bf x}_{t})}\left[\log p_\theta({\bf x}_{0}|{\bf x}_1)\right]\\
    =&\mathbb{E}_{p_\theta({\bf x}_{1}|{\bf x}_{t})}\left[\log\mathcal{N}\left(\frac{1}{\sqrt{1-\beta_1}}\left({\bf x}_1-\frac{\beta_1}{\sqrt{1-\alpha_1^2}}\beps_\theta({\bf x}_1,\alpha_1),\beta_1\mathbf{I}\right)\right)\right]\\
    =&\mathbb{E}_{p_\theta({\bf x}_{1}|{\bf x}_{t})}\left[\frac{D}{2}\log 2\pi\beta_1+\frac{1}{2\beta_1}\left\lVert {\bf x}_0-\frac{1}{\sqrt{1-\beta_1}}\left({\bf x}_1-\frac{\beta_1}{\sqrt{\beta_1}}\beps_\theta({\bf x}_1,\alpha_1)\right)\right\rVert_2^2\right]\\
    =&\frac{D}{2}\log 2\pi\beta_1+\frac{1}{2\beta_1}\mathbb{E}_{p_\theta({\bf x}_{1}|{\bf x}_{t})}\left[\left\lVert \frac{{\bf x}_1-\sqrt{\beta_1}\beps_1}{\sqrt{1-\beta_1}}-\frac{{\bf x}_1-\sqrt{\beta_1}\beps_\theta({\bf x}_1,\alpha_1)}{\sqrt{1-\beta_1}}\right\rVert_2^2\right]\\
    =&\frac{1}{2(1-\beta_{1})}\lVert\beps_1-\beps_\theta\left({\bf x}_1, \alpha_1\right)\rVert_2^2+\frac{D}{2}\log 2\pi\beta_1.
\end{align}

Let $\theta^*=\text{argmin}_\theta \cL^{(t)}_\text{ddpm}(\theta)$ such that $\beps_{\theta^*}\left({\bf x}_t, \alpha_t\right)=\beps_t+{\Delta \beps_t}$ for all $t\in \{1, ..., T\}$, where ${\Delta \beps_t}$ denotes the minimum error vector that can be achieved by $\theta^*$ given the score network $\beps_{\theta}$.
By substituting $\theta^*$ into $\cF^{(t)}_\text{score}(\theta)$ for $t\in \{2, ..., T\}$, we can see that
\begin{align}
    \cF^{(t)}_\text{score}(\theta^*):=&-\mathbb{E}_{q_{\bbeta}(\bx_t|\bx_0)} \left[\mathcal{L}^{(t)}_\text{score}(\theta^*)+\mathcal{R}_{\theta^*}(\bx_0, \bx_t)\right]\\
    =&-\mathbb{E}_{q_{\bbeta}(\bx_t|\bx_0)} \left[\frac{\beta_{t}}{2(1-\beta_{t}-\alpha_{t}^2)}\lVert{\Delta \beps_t}\rVert_2^2+\frac{1}{2(1-\beta_{1})}\lVert{\Delta \beps_1}\rVert_2^2+\frac{D}{2}\log 2\pi\beta_1\right]\\
    =&-\mathbb{E}_{q_{\bbeta}(\bx_t|\bx_0)} \left[\text{min}_\theta\mathcal{L}^{(t)}_\text{score}(\theta)+\text{min}_\theta \mathcal{R}_\theta(\bx_0, \bx_t)\right]\\
    =&\text{max}_\theta\cF^{(t)}_\text{score}(\theta)
\end{align}
\end{proof}

\begin{remark}
Suppose the noise schedule for sampling is monotonic, i.e., $0<\hat{\beta}_{1}<\dotsc < \hat{\beta}_{N}<1$, then, for $1\leq n < N$, $\hat{\beta}_{n}$ satisfies the following inequality:
\begin{align}
     0 < \hat{\beta}_{n} < \min\left\{1 - \frac{\hat{\alpha}_{n+1}^2}{1-\hat{\beta}_{n+1}}, \hat{\beta}_{n+1}\right\}.
\end{align}
\end{remark}

\begin{proof}
By the definition of noise schedule, we know that $0<\hat{\beta}_{1},\dotsc,\hat{\beta}_{N}<1$. Given that $\hat{\alpha}_n=\prod_{i=1}^{n}\sqrt{1-\hat{\beta}_{i}}$, we also have $0<\hat{\alpha}_1,\dotsc,\hat{\alpha}_t<1$. First, we show that $\hat{\beta}_{n} < 1 - \frac{\hat{\alpha}_{n+1}^2}{1-\hat{\beta}_{n+1}}$:
\begin{align}
\hat{\alpha}_{n-1}=\frac{\hat{\alpha}_{n}}{\sqrt{1-\hat{\beta}_{n}}}<1 \Longleftrightarrow
\hat{\beta}_{n} < 1-\hat{\alpha}_{n}^2=1 - \frac{\hat{\alpha}_{n+1}^2}{1-\hat{\beta}_{n+1}}.
\end{align}
Next, we show that $\hat{\beta}_{n}<1-\hat{\alpha}_{n+1}$:
\begin{align}
\frac{\hat{\alpha}_{n}}{\sqrt{1-\hat{\beta}_{n}}}=\frac{\hat{\alpha}_{n}\sqrt{1-\hat{\beta}_{n}}}{1-\hat{\beta}_{n}}=\frac{\hat{\alpha}_{n+1}}{1-\hat{\beta}_{n}}<1 \Longleftrightarrow
\hat{\beta}_{n} < 1-\hat{\alpha}_{n+1}.
\end{align}
Now, we have $\hat{\beta}_{n}<\min\left\{1-\frac{\hat{\alpha}_{n+1}^2}{1-\hat{\beta}_{n+1}}, 1-\hat{\alpha}_{n+1}\right\}$.
In the case of $1 - \hat{\alpha}_{n+1} < 1 - \hat{\alpha}_{n}^2$, we can show that $\hat{\beta}_{n+1} < 1 - \hat{\alpha}_{n+1}$:
\begin{align}
    1-\hat{\alpha}_{n+1}< 1-\hat{\alpha}_{n}^2\Longleftrightarrow
    \hat{\alpha}_{n+1}> \hat{\alpha}_{n}^2\Longleftrightarrow
    \frac{\hat{\alpha}_{n+1}^2}{\hat{\alpha}_{n}^2}> \hat{\alpha}_{n+1}\\\Longleftrightarrow
    1-\frac{\hat{\alpha}_{n+1}^2}{\hat{\alpha}_{n}^2}< 1-\hat{\alpha}_{n+1}\Longleftrightarrow
    \hat{\beta}_{n+1}< 1-\hat{\alpha}_{n+1}.
\end{align}
Therefore, we have $\hat{\beta}_{n}<\min\left\{1-\frac{\hat{\alpha}_{n+1}^2}{1-\hat{\beta}_{n+1}}, \hat{\beta}_{n+1}\right\}$.
\end{proof}

\begin{proposition}
Assuming $\theta$ has been optimized and hypothetically converged to the optimal parameters $\theta^*$, where by optimal parameters it means that $p_{\theta^*}(\bx_{1:t-1}|\bx_0)=q_{\bbeta}(\bx_{1:t-1}|\bx_0)$. Then, we can express the gap between $\log p_{\theta}(\bx_0)$ and $\cF_\text{score}^{(t)}(\theta^{*})$ by $\phi$ as follows:
\begin{align}
\log p_{\theta^{*}}(\bx_0)-\cF_\text{score}^{(t)}(\theta^{*})= \mathbb{E}_{q_{\bbeta}(\bx_{t}|\bx_{0})}\left[\sum_{i=2}^{t}\cL_\text{step}^{(i)}(\phi;\theta^{*})\right],
\end{align}
where
\begin{align}
\cL_\text{step}^{(i)}(\phi;\theta^{*}):=\mathbb{KL}\left(p_{\theta^*} (\bx_{i-1}|\bx_i) || q_\phi (\bx_{i-1}|\bx_0)\right).
\end{align}
\end{proposition}

\begin{proof}
We have
\begin{align}
\log p_{\theta^{*}} ({\bf x}_0) - \mathcal{F}^{(t)}_\text{score}(\theta^{*})
&=\log p_{\theta^{*}} ({\bf x}_0)-\mathbb{E}_{q_{\boldsymbol\beta}({\bf x}_t|{\bf x}_0)}\left[\mathbb{E}_{p_{\theta^{*}}({\bf x}_{1:t-1}|{\bf x}_t)}\left[\log\frac{p_{\theta^{*}}({\bf x}_{0:t-1})}{p_{\theta^{*}}({\bf x}_{1:t-1}|{\bf x}_t)}\right]\right]\\
&=\mathbb{E}_{q_{\boldsymbol\beta}({\bf x}_t|{\bf x}_0)}\left[\mathbb{E}_{p_{\theta^{*}}({\bf x}_{1:t-1}|{\bf x}_t)}\left[\log \frac{p_{\theta^{*}} ({\bf x}_{1:t-1}|{\bf x}_t)}{p_{\theta^{*}} ({\bf x}_{1:t-1}|{\bf x}_0)}\right]\right]\\
&=\mathbb{E}_{q_{\boldsymbol\beta}({\bf x}_t|{\bf x}_0)}\left[\mathbb{E}_{p_{\theta^*}({\bf x}_{1:t-1}|{\bf x}_t)}\left[\log \frac{p_{\theta^*} ({\bf x}_{1:t-1}|{\bf x}_t)}{q_{\bbeta} ({\bf x}_{1:t-1}|{\bf x}_0)}\right]\right]\\
&=\mathbb{E}_{q_{\boldsymbol\beta}({\bf x}_t|{\bf x}_0)}\left[\mathbb{KL}\left(p_{\theta^*} ({\bf x}_{1:t-1}|{\bf x}_t) || q_{\bbeta} ({\bf x}_{1:t-1}|{\bf x}_0)\right)\right]\\
&=\mathbb{E}_{q_{\bbeta}({\bf x}_t|{\bf x}_0)}\left[\sum_{i=2}^{t}\cL_\text{step}^{(i)}(\phi;\theta^{*})\right].
\end{align}
Note that the last equality holds as we considered $q_\phi(\bx_n|\bx_{n-1})=q_{\bbeta}(\bx_t|\bx_{t-1})$.
\end{proof}

\begin{corollary}
Relative to the standard ELBO evaluated at step $t\in\{2, ..., T\}$ defined in \cite{jonathan2020} as
\begin{align}
    \mathcal{F}^{(t)}_\text{elbo}(\theta)&:=-\mathbb{E}_{q_{\bbeta}(\bx_t|\bx_0)}\left[\mathbb{KL}\left(q_{\bbeta}(\mathbf{x}_{t-1}|\mathbf{x}_{t}, \mathbf{x}_{0})||p_{\theta}(\mathbf{x}_{t-1}|\mathbf{x}_{t})\right)+\mathcal{R}_\theta(\bx_0, \bx_t)\right],
\end{align}
we propose a new lower bound as the following
\begin{align}
    \mathcal{F}^{(t)}_\text{bddm}(\theta, \phi)&:=\begin{cases}
        \cF_\text{score}^{(t)}(\theta) \quad & \text{if} \quad \theta \neq \theta^{*}\\
        \cF_\text{score}^{(t)}(\theta)+\mathbb{E}_{q_{\bbeta}(\bx_t|\bx_0)}\left[\cL_\text{step}^{(t)}(\phi;\theta)\right] \quad & \text{if} \quad \theta = \theta^{*}
    \end{cases}
\end{align}
which leads to a tighter lower bound when the conditions in Proposition 2 and 3 are satisfied:
\begin{align}
\log p_{\theta}(\mathbf{x}_0)\geq
\mathcal{F}^{(t)}_\text{bddm}(\theta, \phi)\geq \mathcal{F}^{(t)}_\text{elbo}(\theta).
\end{align}
\end{corollary}
\begin{proof}
For the case of $\theta\neq\theta^*$, the left inequality can be directly justified by the Proposition 1. We can prove the right inequality by Proposition 2, which states that $-\cL^{(t)}_\text{score}(\theta)=\mathcal{F}^{(t)}_\text{elbo}(\theta)$. For the case of $\theta=\theta^*$, by Proposition 3, we have $\log p_{\theta^*}({\bf x}_0)=\mathcal{F}^{(t)}_\text{score}(\theta^{*})+\sum_{i=2}^{t}\cL_\text{step}^{(i)}(\phi;\theta^{*})=\mathcal{F}^{(t)}_\text{bddm}(\theta^{*}, \phi)+\sum_{i=2}^{t-1}\cL_\text{step}^{(i)}(\phi;\theta^{*})\geq\mathcal{F}^{(t)}_\text{bddm}(\theta^{*}, \phi)\geq\mathcal{F}^{(t)}_\text{elbo}(\theta^{*})$, since $\cL_\text{step}^{(i)}(\phi;\theta^{*})\geq 0$ $\forall i$.
\end{proof}
 
\begin{note}
\begin{align}
    \cL_\text{step}^{(t)}(\phi;\theta^*)&=\frac{1}{2(1-\hat{\beta}_t(\phi)-{\alpha}_{t}^2)}\left\lVert\sqrt{1-\alpha_t^2}\beps_t - \frac{\hat{\beta}_t(\phi)}{\sqrt{1-\alpha_t^2}}\beps_{\theta^*}(\bx_t, \alpha_t) \right\rVert^2_2+C_t(\phi),
\end{align}
where
\begin{align}
    C_t(\phi):=\frac{1}{4}\log \frac{1-{\alpha}^2_t}{\hat{\beta}_t(\phi)}+\frac{D}{2}\left(\frac{\hat{\beta}_t(\phi)}{1-{\alpha}_t^2}-1\right).
\end{align}
\end{note}
\begin{proof}
By the definition of the following probability density functions:
\begin{align}
    p_\theta(\bx_{t-1}|\bx_{t})&
    =
    \cN
    \left(
        \frac
        {1}{\sqrt{1-\beta_{t}}}
        \left(
        \bx_{t}
        -
        \frac
        {\beta_{t}}
        {\sqrt{1-\alpha_{t}^2}}
        \beps_\theta\left(\bx_{t}, \alpha_{t}\right)
        \right),
        \frac
        {(1-\beta_t-\alpha_{t}^2)\beta_{t}}
        {(1-\beta_t)(1-\alpha_{t}^2)}
        \bI
    \right)\\
    q_\phi(\bx_{t-1}|\bx_{0})&
    =
    \cN
    \left(
        \alpha_{t-1}\bx_{0},\left(1-\alpha_{t-1}^2\right) \bI
    \right)=\cN
    \left(
        \frac{\alpha_{t}}{\sqrt{1-\beta_{t}}}\bx_{0},\left(1-\frac{\alpha_{t}^2}{1-\beta_{t}}\right) \bI
    \right),
\end{align}
we have
\begin{align}
    &\cL_\text{step}^{(t)}(\phi;\theta^*):=\KL\left(p_{\theta^*}(\bx_{t-1}|\bx_{t})||q_\phi(\bx_{t-1}|\bx_0)\right)\\
    =
    &\frac{1-\beta_t}{2(1-\beta_t-{\alpha}_{t}^2)}
    \left\lVert
        \frac{\alpha_{t}}{\sqrt{1-\beta_{t}}}\bx_{0}
        -
        \frac
        {1}{\sqrt{1-\beta_{t}}}
        \left(
        \bx_{t}
        -
        \frac
        {\beta_{t}}
        {\sqrt{1-\alpha_{t}^2}}
        \beps_{\theta^*}\left(
            \bx_{t}, \alpha_{t}
        \right)
    \right)
    \right\rVert_{2}^{2}
    +C_t\\
    =
    &\frac{1-\beta_t}{2(1-\beta_t-{\alpha}_{t}^2)}
    \left\lVert
        \frac{\alpha_{t}}{\sqrt{1-\beta_{t}}}\bx_{0}
        -
        \frac
        {1}{\sqrt{1-\beta_{t}}}
        \left(
        \alpha_{t}\bx_{0} + \sqrt{1-\alpha_{t}^2}\beps_t
        -
        \frac
        {\beta_{t}}
        {\sqrt{1-\alpha_{t}^2}}
        \beps_{\theta^*}\left(
            \bx_{t}, \alpha_{t}
        \right)
    \right)
    \right\rVert_{2}^{2}
    +C_t\\
    =
    &\frac{1-\beta_t}{2(1-\beta_t-{\alpha}_{t}^2)}\left\lVert\sqrt{\frac{1-\alpha_t^2}{1-{\beta}_t}}\beps_t - \frac{\beta_t}{\sqrt{(1-{\beta}_t)(1-\alpha_t^2)}}\beps_{\theta^*}(\bx_t, \alpha_t) \right\rVert^2_2+C_t\\
    =
    &\frac{1}{2(1-\beta_t-{\alpha}_{t}^2)}\left\lVert\sqrt{1-\alpha_t^2}\beps_t - \frac{\beta_t}{\sqrt{1-\alpha_t^2}}\beps_{\theta^*}(\bx_t, \alpha_t) \right\rVert^2_2+C_t,
\end{align}
where
\begin{align}
    C_t=\frac{1}{4}\log \frac{1-{\alpha}^2_t}{{\beta}_t}+\frac{D}{2}\left(\frac{{\beta}_t}{1-{\alpha}_t^2}-1\right).
\end{align}
As we parameterize $\beta_t$ by $\phi$ to estimate the noise schedule for inference, we obtain the final step loss:
\begin{align}
    \cL_\text{step}^{(t)}(\phi;\theta^*)&=\frac{1}{2(1-\hat{\beta}_t(\phi)-{\alpha}_{t}^2)}\left\lVert\sqrt{1-\alpha_t^2}\beps_t - \frac{\hat{\beta}_t(\phi)}{\sqrt{1-\alpha_t^2}}\beps_{\theta^*}(\bx_t, \alpha_t) \right\rVert^2_2+C_t(\phi),
\end{align}
where
\begin{align}
    C_t(\phi):=\frac{1}{4}\log \frac{1-{\alpha}^2_t}{\hat{\beta}_t(\phi)}+\frac{D}{2}\left(\frac{\hat{\beta}_t(\phi)}{1-{\alpha}_t^2}-1\right).
\end{align}
\end{proof}

\section{Experimental details}
\subsection{Conventional grid search algorithm for DDPMs}
\label{sec:gridsearch}
We reproduced the grid search algorithm in \cite{nanxin2020}, in which a 6-step noise schedule was searched. In our paper, we generalized the grid search algorithm by similarly sweeping the $N$-step noise schedule over the following possibilities with a bin width $M=9$:
\begin{align}
    \{1, 2, 3, 4, 5, 6, 7, 8, 9\} \otimes \{10^{-6\cdot N/N}, 10^{-6\cdot (N-1)/N}, ..., 10^{-6\cdot 1/N}\},
\end{align}
where $\otimes$ denotes the cartesian product applied on two sets. LS-MSE was used as a metric to select the solution during the search. When $N=6$, we resemble the GS algorithm in \cite{nanxin2020}. Note that above searching method normally does not scale up to $N>8$ steps for its exponential computational cost $\mathcal{O}(9^{N})$.

\subsection{Hyperparameter setting in BDDMs}
Algorithm \ref{alg:training_phi} took a skip factor $\tau$ to control the stride for training the scheduling network. The value of $\tau$ would affect the coverage of step sizes when training the scheduling network, hence affecting the predicted number of steps $S$ for inference -- the higher $\tau$ is, the shorter $S$ tends to be. We set $\tau=200$ and $\tau=20$ for speech synthesis and image generation, respectively.

For initializing Algorithm \ref{alg:nspred} for noise scheduling, we could take as few as $1$ training sample for validation, perform a grid search on the hyperparameters $\{(\alpha_N=0.1i, \beta_N=0.1j)\}$ for $i,j=1,...,9$, i.e., $81$ possibilities in all, and use the LS-MSE measure as the selection metric. Then, the predicted noise schedule corresponding to the least LS-MSE was stored and applied to the online inference afterward, as shown in Algorithm \ref{alg:sampling}. Note that this searching has a complexity of only $\mathcal{O}(M^{2})$ (e.g., $M=9$ in this case), which is much more efficient than $\mathcal{O}(M^{N})$ in the conventional grid search algorithm in \cite{nanxin2020}, as discussed in Section \ref{sec:gridsearch}.


\begin{table}[t]
\centering
\caption{Performances of different noise schedules on the multi-speaker VCTK speech dataset, each of which used the same score network \cite{nanxin2020} $\beps_{\theta}(\cdot)$ that was trained on VCTK for about 1M iterations.
}
\label{tab:vctk}
\begin{tabular}{lccccc}
 \toprule
        {\bfseries Noise schedule} & \bfseries LS-MSE ($\downarrow$) & \bfseries MCD ($\downarrow$) &\bfseries STOI ($\uparrow$) &\bfseries PESQ ($\uparrow$) & \bfseries MOS ($\uparrow$) \\
 \midrule
 \multicolumn{6}{l}{\bf DDPM \cite{jonathan2020, nanxin2020}} \\
 \quad 8 steps (Grid Search) & 101 & \bf 2.09 & \bf 0.787 & \bf 3.31 & \bf 4.22 $\pm$ 0.04\\
 \quad 1,000 steps (Linear) & 85.0 & 2.02 & 0.798 & 3.39 & 4.40 $\pm$ 0.05\\
 \midrule
 \multicolumn{6}{l}{\bf DDIM \cite{jiaming2021}} \\
 \quad 8 steps (Linear) & 553 & 3.20 & 0.701 & 2.81 & 3.83 $\pm$ 0.04\\
 \quad 16 steps (Linear) & 412 & 2.90 & 0.724 & 3.04 & 3.88 $\pm$ 0.05\\
 \quad 21 steps (Linear) & 355 & 2.79 & 0.739 & 3.12 & 4.12 $\pm$ 0.05\\
 \quad 100 steps (Linear) & 259 & 2.58 & 0.759 & 3.30 & 4.27 $\pm$ 0.04\\
 \midrule
 \multicolumn{6}{l}{\bf NE \cite{san2021noise}} \\
 \quad 8 steps (Linear) & 208 & 2.54 & 0.740 & 3.10 & 4.18 $\pm$ 0.04\\ 
 \quad 16 steps (Linear) & 183 & 2.53 & 0.742 & 3.20 & 4.26 $\pm$ 0.04\\
 \quad 21 steps (Linear) & 852 & 3.57 & 0.699 & 2.66 & 3.70 $\pm$ 0.03\\
 \midrule
 \multicolumn{6}{l}{\bf BDDM $(\hat{\alpha}_N,\hat{\beta}_N)$} \\
 \quad 8 steps $(0.2, 0.9)$ & \bf 98.4 & 2.11 & 0.774 & 3.18 & 4.20 $\pm$ 0.04 \\
 \quad 16 steps $(0.5, 0.5)$ & \bf 73.6 & \bf 1.93 & \bf 0.813 & \bf 3.39 & \bf 4.35 $\pm$ 0.05\\
 \quad 21 steps $(0.5, 0.1)$ & \bf 76.5 & \bf 1.83 & \bf 0.827 & \bf 3.43 & \bf 4.48 $\pm$ 0.06 \\
 \bottomrule
\end{tabular}
\end{table}
\section{Additional experiments}
A demonstration page at \textcolor{blue}{\url{https://bilateral-denoising-diffusion-model.github.io}} shows some samples generated by BDDMs trained on LJ speech, VCTK, and CIFAR-10\footnote{A detailed discussion on image generation using BDDMs is coming soon.} datasets.

\subsection{Multi-speaker speech synthesis}
In addition to the single-speaker speech synthesis, we evaluated BDDMs on the multi-speaker speech synthesis benchmark VCTK \cite{yamagishi2019vctk}. VCTK consists of utterances sampled at $48$ KHz by $108$ native English speakers with various accents. 
We split the VCTK dataset for training and testing: 100 speakers were used for training the multi-speaker model and 8 speakers for testing.
We trained on a 44257-utterance subset (40 hours) and evaluated on a held-out 100-utterance subset. During training, similar to Wavegrad \cite{nanxin2020}, mel-spectrograms computed from ground-truth audio were used as the conditioning features. In addition, we used the held-out subset for evaluating synthesized speech with the ground-truth features. 

Results are presented in Table \ref{tab:vctk}. For this multi-speaker VCTK dataset, we obtained consistent observations with that for the single-speaker LJ dataset.
Again, the proposed BDDM with only 16 or 21 steps outperformed the DDPM with 1,000 steps. To the best of our knowledge, ours was the first work that reported this degree of superior. When reducing to 8 steps, BDDM obtained performance on par with (except for a worse PESQ) the costly grid-searched 8 steps (which were unscalable to more steps) in DDPM.  
For NE, we could again observe a degradation from its 16 steps to 21 steps, indicating the instability of NE for the VCTK dataset likewise. In contrast, BDDM gave continuously improved performance while increasing the step number.

\pgfplotscreateplotcyclelist{colorlist}{%
blue,every mark/.append style={fill=blue},mark=*\\%
red,every mark/.append style={fill=red},mark=*\\%
green!80!black,every mark/.append style={fill=green!80!black},mark=*\\%
brown,every mark/.append style={fill=brown},mark=*\\%
pink,every mark/.append style={fill=pink},mark=*\\%
cyan,every mark/.append style={fill=cyan},mark=*\\%
green!60!black,densely dashed,every mark/.append style={
solid,fill=brown!100!black},mark=otimes*\\%
brown,densely dashed,every mark/.append style={solid,fill=gray},mark=otimes*\\%
blue,densely dashed,mark=star,every mark/.append style=solid\\%
red,densely dashed,every mark/.append style={solid,fill=red!80!black},mark=diamond*\\%
}

\begin{figure}[t]
\begin{minipage}[H]{0.49\textwidth}
\begin{figure}[H]
    \centering
\resizebox{\columnwidth}{!}
{
\includegraphics{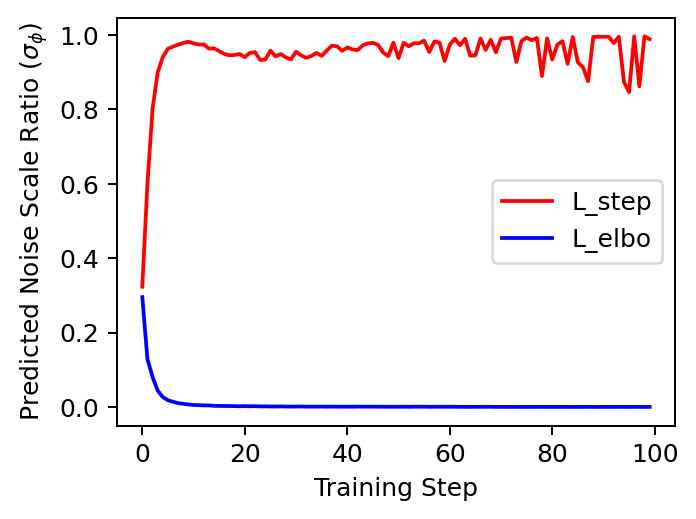}
}
    \caption{Different training losses for $\sigma_\phi$}\label{fig:step_loss}
\end{figure}
\end{minipage}
\begin{minipage}[H]{0.49\textwidth}
\begin{figure}[H]
    \centering
\resizebox{\columnwidth}{!}
{
\includegraphics{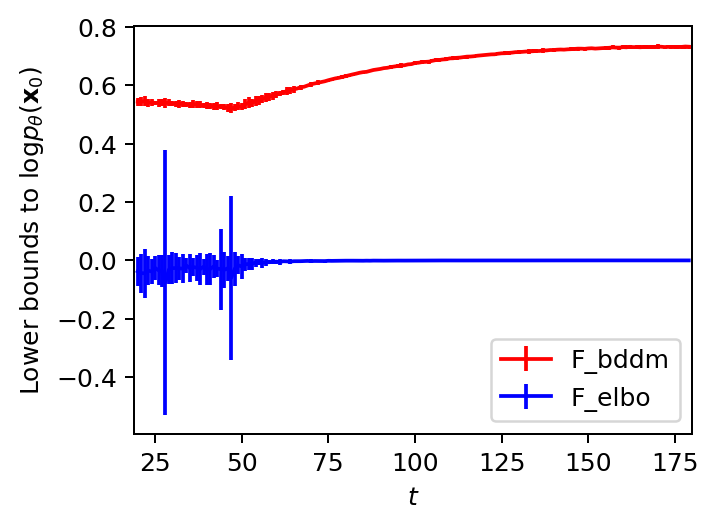}
    }
    \caption{Different lower bounds to $\log p_\theta(\bx_0)$}\label{fig:bounds}
\end{figure}
\end{minipage}
\end{figure}

\subsection{Comparing against ELBO for training the scheduling network}

We also conducted an experiment to investigate the usefulness of $\cF_\text{elbo}^{(t)}$ in learning $\phi$. Similar to the derivation of $\cL_\text{step}^{(t)}$, we reparameterize $\beta_t$ in $\cF_\text{elbo}^{(t)}$ by $\hat{\beta}_t(\phi)$ and used it to train the scheduling network $\sigma_\phi$. Figure \ref{fig:step_loss} compares the scheduling networks' outputs when training with $\cL^{(t)}_\text{step}$ and $\cL^{(t)}_\text{elbo}$, respectively, on the same LJ training dataset. The plot shows that when using $\cL^{(t)}_\text{elbo}$ to learn $\phi$, the network output rapidly collapsed to zero within several training steps; whereas, the network trained with $\cL^{(t)}_\text{step}$ produced outputs fluctuating around 1. The fluctuation is a desirable property indicating that the network can predict a $t$-dependent noise scale, where $t$ is a random time step drawn from an uniform distribution.

\subsection{Comparing against ELBO for lower bounding log evidence}

To validate the inequality of lower bounds: $\mathcal{F}^{(t)}_\text{bddm}(\theta, \phi) \geq \mathcal{F}^{(t)}_\text{elbo}(\theta)$, we evaluated their respective values at different time step $t$ using the same optimized set of parameters $(\theta^{*}, \phi^{*})$. The result is illustrated in Figure \ref{fig:bounds}, where each value is provided with 95\% confidence intervals. Notably, the common entropy term $\E_{q_\phi(\bx_t|\bx_0)}\left[\mathcal{R}_\theta(\bx_0, \bx_t)\right] < 0$ was dropped to mainly compare the KL terms. Therefore, the plotted lower bound values might be positive. The graph shows that the proposed new bound $\mathcal{F}^{(t)}_\text{bddm}$ is always better than the standard one across all examined $t$. Interestingly, we found that $\mathcal{F}^{(t)}_\text{elbo}$ became highly varied when $t\in [20, 50]$, in contrast, $\mathcal{F}^{(t)}_\text{bddm}$ attained the lowest values at the range $t\in [20, 50]$ with a relatively much lower variance. This reveals the superiority of $\mathcal{F}^{(t)}_\text{bddm}$ that its absolute value can better correspond to the difficulty of training, especially when $t$ is close to zero.

\subsection{Comparing different reverse processes for BDDMs}
This section demonstrates that BDDMs do not restrict the sampling procedure to a specialized reverse process in Algorithm \ref{alg:sampling}. In particular, we evaluated different reverse processes, including that of DDPMs (see Eq. \ref{eq:reverse}) and DDIMs \cite{jiaming2021}, for BDDMs and compared the objective scores on the generated samples. DDIMs \cite{jiaming2021} formulate a non-Markovian generative process that accelerates the inference while keeping the same training procedure as DDPMs. The original generative process in Eq. \ref{eq:reverse} in DDPMs is modified into
\begin{align}
\label{eq:ddim}
    p_\theta^{(\tau)}(\bx_{0:T})
    :=
    \pi(\bx_T)\prod_{i=1}^Sp_\theta^{(\gamma_i)}(\bx_{\gamma_{i-1}}|\bx_{\gamma_i})
    \times
    \prod_{t\in{\bar{\gamma}}}p_\theta^{(t)}(\bx_0|\bx_t),
\end{align}
where ${\boldsymbol\gamma}$ is a sub-sequence of length $N$ of $[1,...,T]$ with $\gamma_N=T$, and $\bar{{\boldsymbol\gamma}}:=\{1,...,T\}\setminus{\boldsymbol\gamma}$ is defined as its complement; Therefore, only part of the models are used in the sampling process. 

To achieve the above, DDIMs defined a prediction function $f_{\theta}^{(t)}(\bx_t)$ that depends on $\beps_{\theta}$ to predict the observation $\bx_0$ given $\bx_t$ directly:
\begin{align}
\label{eq:pf}
    f_{\theta}^{(t)}(\bx_t)&:=\frac{1}{\alpha_t}\left(\bx_t-\sqrt{1-\alpha_t^2}\beps_{\theta}(\bx_t,\alpha_t)\right).
\end{align}
By leveraging this prediction function, the conditionals in Eq. \ref{eq:ddim} are formulated as
\begin{align}
\label{eq:ddimreverse}
    p_\theta^{(\gamma_i)}(\bx_{\gamma_{i-1}}|\bx_{\gamma_i}) 
    &=
     \cN
    \left(
    \frac{\alpha_{\gamma_{i-1}}}{\alpha_{\gamma_{i}}}
    \left(
    \bx_{\gamma_i}-
    \varsigma
    \beps_{\theta}(\bx_{\gamma_i},\alpha_{\gamma_i})
    \right),\sigma_{\gamma_i}^2\bI
    \right)\quad \text{if }\! i\in[N], i>1\\
\label{eq:ddimreverse2}
    p_\theta^{(t)}(\bx_0|\bx_t)
    &=\cN(f_{\theta}^{(t)}(\bx_t),\sigma_t^2\bI)\quad \text{otherwise,}
\end{align}
where the detailed derivation of $\sigma_t$ and $\varsigma$ can be referred to \cite{jiaming2021}. In the original DDIMs, the accelerated reverse process produces samples over the subsequence of $\bbeta$ indexed by ${\boldsymbol\gamma}$: $\hat{\bbeta}=\{\beta_n|n\in{\boldsymbol\gamma}\}$. In BDDMs, to apply the DDIM reverse process, we use the $\hat{\bbeta}$ predicted by the scheduling network in place of a subsequence of the training schedule $\bbeta$.

Finally. the objective scores are given in Table \ref{tab:lj_reverse_comp}. Note that the subjective evaluation (MOS) is omitted here since the other assessments above have shown that the MOS scores are highly correlated with the objective measures, including STOI and PESQ. They indicate that applying BDDMs to either DDPM or DDIM reverse process leads to comparable and competitive results. Meanwhile, the results show some subtle differences: BDDMs over a DDPM reverse process gave slightly better samples in terms of signal error and consistency metrics (i.e., LS-MSE and MCD), while BDDM over a DDIM reverse process tended to generate better samples in terms of intelligibility and perceptual metrics (i.e., STOI and PESQ).
\begin{table}[t]
\centering
\caption{Performances of different reverse processes for BDDMs on the LJ speech dataset, each of which used the same score network \cite{nanxin2020} $\beps_{\theta}(\cdot)$ and the same noise schedule.
}
\label{tab:lj_reverse_comp}
\begin{tabular}{lcccc}
 \toprule
        {\bfseries Noise schedule} & \bfseries LS-MSE ($\downarrow$) & \bfseries MCD ($\downarrow$) &\bfseries STOI ($\uparrow$) &\bfseries PESQ ($\uparrow$) \\
 \midrule
 \multicolumn{5}{l}{\bf BDDM (DDPM reverse process)} \\
 \quad 8 steps $(0.3, 0.9, 1e^{-5})$ & \bf 91.3 & \bf 2.19 & 0.936 & 3.22\\
 \quad 16 steps $(0.7, 0.1, 1e^{-6})$ & \bf 73.3 & \bf 1.88 & 0.949 & 3.32\\
 \quad 21 steps $(0.5, 0.1, 1e^{-6})$ & \bf 72.2 & \bf 1.91 & 0.950 & 3.33\\
 \midrule
 \multicolumn{5}{l}{\bf BDDM (DDIM reverse process)} \\
 \quad 8 steps $(0.3, 0.9, 1e^{-5})$ & 91.8 & \bf 2.19 & \bf 0.938 & \bf 3.26\\
 \quad 16 steps $(0.7, 0.1, 1e^{-6})$ & 77.7 & 1.96 & \bf 0.953 & \bf 3.37\\
 \quad 21 steps $(0.5, 0.1, 1e^{-6})$ & 77.6 & 1.96 & \bf 0.954 & \bf 3.39\\
 \bottomrule
\end{tabular}
\end{table}

\subsection{Implementation details}
Our proposed BDDMs and the baseline methods were all implemented with the Pytorch library. The score networks for the LJ and VCTK speech datasets were trained from scratch on a single NVIDIA Tesla P40 GPU with batch size $32$ for about 1M steps, which took about 3 days. In comparison, the training of scheduling networks for BDDMs took only 10k steps to converge, which consumed no more than an hour for all three datasets. More details regarding the model architecture, the Pytorch implementation, and the type of resources used can be found in our code provided in the supplementary materials.


\subsection{Crowd-sourced subjective evaluation}
All our Mean Opinion Score (MOS) tests were crowd-sourced. We refer to the MOS scores in~\cite{protasio_ribeiro_crowdmos_2011}, and the scoring criteria have been included in Table~\ref{matrix:naturalness} for completeness. The samples were presented and rated one at a time by the testers.

\begin{table}[H]
  \caption{Ratings that have been used in evaluation of speech naturalness of synthetic samples.}
  \label{matrix:naturalness}
  \begin{tabular}{ccc}
  \toprule
  Rating & Naturalness & Definition                           \\
  \midrule
  1      & Unsatisfactory        &  Very annoying, distortion is objectionable. \\
  2      & Poor       &  Annoying distortion, but not objectionable. \\
  3      & Fair       &  Perceptible distortion, slightly annoying.\\
  4      & Good       & Slight perceptible level of distortion, but not annoying.\\
  5      & Excellent  & Imperceptible level of distortion.\\
  \bottomrule
  \end{tabular}
  \end{table}

\newpage
\begin{figure}[H]
    \centering
    \includegraphics[scale=1]{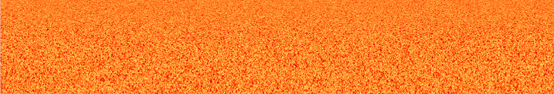}
    \includegraphics[scale=1]{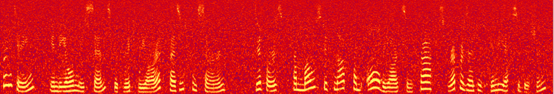}
    \includegraphics[scale=1]{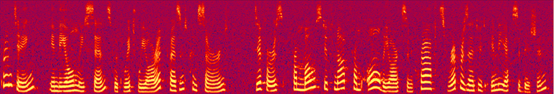}
    \includegraphics[scale=1]{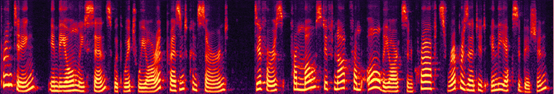}
    \caption{Spectrum plots of the speech samples produced by a well-trained BDDM within 3 sampling steps. The first row shows the spectrum of a random signal for starting the reverse process. Then, from the top to the bottom, we show the spectrum of the resultant signal after each step of the reverse process performed by the BDDM. We also provide the corresponding WAV files on our project page.}
\end{figure}

\end{document}